\documentclass[preprint,12pt,3p]{elsarticle}

\usepackage{amsmath,amssymb,amsthm}
\usepackage{graphicx}
\usepackage{booktabs}
\usepackage{array}
\usepackage{longtable}
\usepackage{multirow}
\usepackage{enumitem}
\usepackage{xcolor}
\usepackage{hyperref}
\usepackage{subcaption}
\usepackage{algorithm}
\usepackage{algpseudocode}
\usepackage{siunitx}

\hypersetup{
  colorlinks=true,
  linkcolor=blue,
  citecolor=blue,
  urlcolor=blue,
}

\newtheorem{proposition}{Proposition}

\newcommand{\Qtarget}{Q^{*}_{\text{target}}}
\newcommand{\ftilde}{\tilde{f}}
\newcommand{\fcorr}{\tilde{f}^{\text{corr}}}
\newcommand{\sbase}{s^{\text{base}}}
\newcommand{\seff}{s^{\text{eff}}}
\newcommand{\Bmin}{B_{\min}}
\newcommand{\Bk}{B_{k}}
\newcommand{\deltak}{\delta_{k}}
\newcommand{\phik}{\phi_{k}}

\newcommand{\fagg}{\bar{f}}                     
\newcommand{\ftgt}{\hat{f}^{\text{tgt}}}        
\newcommand{\gvec}{\mathbf{g}}                  
\newcommand{\hgeo}{\mathbf{h}^{\text{geo}}}     
\newcommand{\xseq}{\mathbf{x}^{\text{seq}}}     
\newcommand{\zseq}{\mathbf{z}^{\text{seq}}}     
\newcommand{\Hconv}[1]{\mathbf{H}^{(#1)}}       
\newcommand{\ddis}{d^{\text{dis}}}              
\newcommand{\Dclean}{D^{\text{dis}}_{\text{clean}}}  
\newcommand{\Dtgtbg}{D_{\text{tgt-bg}}}         
\newcommand{\Cintra}{C_{\text{intra}}}          
\newcommand{\Ewl}{E_{\text{wl}}}                
\newcommand{\Sarea}{S_{\text{area}}}            
\newcommand{\Npatch}{N_{\text{patch}}}          
\newcommand{\wcred}{w^{\text{cred}}}            
\newcommand{\gagg}{g^{\text{agg}}}              
\newcommand{\lqual}{\lambda^{\text{qual}}}      
\newcommand{\bqual}{b^{\text{qual}}}            
\newcommand{\etastu}{\eta^{\text{stu}}}         
\newcommand{\etacons}{\eta^{\text{cons}}}       
\newcommand{\phij}{\phi_j}                      

\journal{Displays}

\begin{document}

\begin{frontmatter}

\title{BiCG-Net: Target-Specific Image Quality Assessment via Observation Modeling and In-Image Differencing}

\author[a]{Yang Yibu}
\author[a]{Cui Tingqiong}
\author[a]{Lei Liming}
\author[a]{Xiong Hao}
\author[b]{Xu Jun}
\author[c]{Liu Siyuan}
\author[b]{Huang Guanghui\texorpdfstring{\corref{cor1}}{}}

\cortext[cor1]{Corresponding author.}
\address[a]{CRRC Chongqing Co., Ltd., Chongqing, China}
\address[b]{College of Mathematics and Statistics, Chongqing University, Chongqing, China}
\address[c]{State Key Laboratory of Mechanical Transmission for Advanced Equipment, Chongqing University, Chongqing, China}

\begin{abstract}
We estimate the quality of a \emph{specific target region} (e.g., a
train body) in an image, rather than the quality of the whole image.
This target-level task differs fundamentally from whole-image
No-Reference Image Quality Assessment (NR-IQA): when the target
occupies only a small fraction of the frame, the whole-image score is
dominated by background, and no public benchmark provides target-level
human Mean Opinion Score (MOS). We therefore estimate target quality
by \emph{nested sampling}---a sequence of patches expanding outward
from the target box---and infer quality from the relative structure of
this sequence rather than from any single patch. This motivates
BiCG-Net, a student network that distills an 11-teacher ensemble
through a linear observation model and an \emph{in-image differencing}
mechanism, which provably cancels the target term and the
teacher-specific bias. We train and test the student independently on
each dataset, and separately evaluate zero-shot transfer across
benchmarks. On CSIQ and LIVEC, where the target-level task degenerates
to whole-image assessment, the student reaches PLCCs of 0.9009 and
0.9133 against human MOS, exceeding the best single teacher and the
teacher consensus on both datasets. On a self-built MOS-free railway
dataset, the student reaches a PLCC of 0.9855 against the weak label,
not human MOS. The student, distilled from these teachers, exceeds its
own teachers because it is not trained to mimic any of them: it learns
a new mapping from the \emph{relative structure} of teacher readings to
the ground-truth label. A residual analysis on CSIQ and LIVEC compares
the student's prediction with the teacher consensus. The residual is
significantly correlated with background contamination, teacher
dispersion, and label extremity, showing that the student extracts
information the consensus does not contain. Under CSIQ-to-LIVEC
transfer, the frozen student reaches a PLCC of 0.8416 against human MOS
on LIVEC, exceeding the consensus by 0.1947. We position BiCG-Net as an
offline target-level quality auditing and label-generation engine. Its
39,482-parameter student is compact by design, acting as a unifying
layer that maps the 11 heterogeneous teacher readings onto a common
quality scale.
\end{abstract}


\end{frontmatter}

\section{Introduction}
\label{sec:intro}

Deploying an image quality assessment (IQA) system in practical vision
pipelines requires estimating the quality of a \emph{specific target
region}, not of the whole image. Consider a railway rolling-stock
inspection system: the train body typically occupies only a small
fraction of the frame, while the rest is background clutter, track
ballast, and vegetation. A whole-image quality score is therefore
dominated by irrelevant background and provides little information
about the target itself. In such settings, the primary objective is
the accuracy of target-level quality estimation: downstream operations
depend on high-fidelity, deterministic quality labels, and a fast but
inaccurate score is of little use. Establishing target-level accuracy
without resorting to expensive, logistically prohibitive target-level
re-annotation campaigns is therefore the core requirement.

This target-level requirement becomes especially stringent when the
quality labels are used for downstream generative-AI training and data
curation. Generative models are sensitive to their training
distributions: training on blurred, specularly reflective, or partially
occluded targets leads to structural artifacts and inherited motion
blur in the synthesized output. The fidelity of target-level quality
labels therefore bounds the synthesis capability of downstream
architectures, making accurate target-level assessment a prerequisite
for offline data curation.

An intuitive baseline is to apply off-the-shelf No-Reference IQA
(NR-IQA) models directly to the local region of interest. However,
contemporary NR-IQA methodologies are predominantly tailored for global,
\emph{whole-image} quality prediction
\cite{wang2004iqa,mao2025nr,ghadiyaram2017perceptual}. Classical
statistical schemes like BRISQUE \cite{mittal2012brisque} and NIQE
\cite{mittal2012niqe} extract natural scene statistics across the entire
spatial domain, whereas deep paradigms such as MUSIQ \cite{ke2021musiq},
MANIQA \cite{yang2022maniqa}, and CLIP-IQA+ \cite{wang2023clip-iqa-plus}
aggregate global representations contextualized via transformer blocks
or vision-language foundation models. These methods operate under the
monolithic assumption that the entire frame constitutes a singular,
homogeneous quality unit, which collapses when the target occupies only
a small fraction of the frame: the aggregated score is then dominated
by background, while the structural integrity of the target is heavily
contaminated. This motivates \emph{nested sampling}: instead of scoring
a single target crop, we score a sequence of nested patches and infer
the target quality from how teacher readings change as the patch
expands outward.

Our goal is to assign a reliable quality score to a specific target in
an image without target-level MOS. This motivates BiCG-Net, which
models each teacher reading as a linear decomposition of five latent
quantities (target quality, teacher bias, area effect, background
content, and noise) and cancels the target term and teacher bias by
\emph{in-image differencing}---subtracting the readings of two patches
in the same image. In the
experiments, the student is trained and tested independently on each
dataset, and evaluated separately under zero-shot cross-dataset
transfer.

Public benchmarks provide only whole-image human MOS, not
target-level MOS. To validate the target-level formulation on these
benchmarks, we deliberately designate a center $1/9$ area as the
pseudo-target and expand the nested sampling until the patch boundary
coincides with the full frame.\footnote{The $1/9$ ratio keeps the
target small relative to the frame while avoiding a degenerate
single-pixel target.} Because the ultimate sampling patch is the whole
image, the target-level evaluation mathematically degenerates to a
global prediction, allowing us to leverage whole-image human MOS to
validate the target-level formulation.

We evaluate on three datasets: two public MOS benchmarks (CSIQ and
LIVEC) and one self-built MOS-free railway dataset. On CSIQ and LIVEC,
where the target-level task degenerates to whole-image assessment, the
student reaches PLCCs of 0.9009 and 0.9133 against human MOS, exceeding
both the best single teacher and the teacher consensus on both
datasets. On the self-built MOS-free railway dataset, the student
reaches a PLCC of 0.9855 against the weak label (a consensus of
teacher readings), not human MOS. Under a strictly zero-shot
CSIQ-to-LIVEC transfer, the frozen student reaches a PLCC of 0.8416,
exceeding the consensus by $+0.1947$; the reverse direction essentially
matches the consensus, an asymmetry we hypothesize to reflect the
information-coverage gap of the source manifold and analyze further
through the $\Delta$ diagnostics in Section~\ref{sec:ana-delta}.

This gap cannot be attributed to network capacity alone; it stems from
the mathematical structure of raw teacher readings. When a whole-image
IQA teacher evaluates an expanded patch that encloses the target, its
scalar output is jointly governed by five latent quantities: the
target quality, the teacher-specific bias, the area effect, the
background content, and measurement noise. Because these components
are individually latent and underdetermined, simple arithmetic
averaging across an ensemble propagates both the teacher bias and the
background contribution into the final supervision signal. When the
target occupies only a small fraction of the frame, these two
components dominate the average, causing consensus networks to output
background-driven scores rather than target-level quality.

The core mathematical insight is that, although the five components
are individually unobserved, the target quality and the
teacher-specific bias remain invariant within a single image, whereas
the area effect and the background content vary across the patch
sequence. Taking the difference between any two patches in the same
image therefore cancels the target term and the teacher bias, leaving
only the area effect, the background difference, and noise. By
compensating for the area effect through a dedicated geometric encoder,
the residual reduces to a pure background-and-noise term, a property
we term \emph{in-image differencing}. Because the operation is
performed within a single image, the resulting statistics are
within-image invariant by construction, free from the absolute scales
and site-specific biases of the teacher models. BiCG-Net builds on
this mechanism to estimate target quality without target-level MOS.

The main contributions of this work are as follows:
\begin{enumerate}
\item \textbf{A nested-sampling formulation with provable bias
      elimination.} We formulate target quality as a relative property
      of a nested patch sequence and prove that in-image differencing
      exactly eliminates the target term and the teacher systematic
      bias, delivering a bias-eliminated supervision signal for
      multi-teacher distillation.
\item \textbf{A formal underdetermined observation model.} We
      formalize raw teacher readings as a linear decomposition of five
      latent variables, and prove that conventional consensus
      aggregation is vulnerable when the target occupies only a small
      fraction of the frame, because unmodeled teacher biases and
      background content leak into the average.
\item \textbf{An interpretable four-module architecture.} We design
      BiCG-Net as a direct mathematical reflection of the observation
      components: per-teacher scaling, observation correction, sequence
      encoding, and gated aggregation. The architecture is compact by
      design, so that it learns to reconcile the 11 heterogeneous
      teacher readings rather than memorize individual teachers.
\item \textbf{A validation across perceptual and deployment tracks
      with a $\Delta$ diagnostic.} We separate perceptual validity
      (MOS benchmarks) from deployment usability (the MOS-free
      dataset), and characterize information beyond the teacher
      ensemble through a $\Delta$ analysis that explains both the
      in-distribution gains and the asymmetric transfer (the
      CSIQ-to-LIVEC gain vs.\ the flat reverse direction).
\end{enumerate}
\section{Related Work}
\label{sec:related}

\subsection{Whole-Image No-Reference Image Quality Assessment}
\label{sec:rw-whole}

No-Reference Image Quality Assessment (NR-IQA) aims to predict the
perceptual quality of an image without access to a pristine reference.
The field has evolved through three distinct generations, which we
review below; a comprehensive survey of DNN-based methods can be found
in \cite{yang2019survey}.

\noindent\textbf{Hand-crafted statistics.} The first generation relies
on the theory of Natural Scene Statistics (NSS), assuming that natural
images occupy a characteristic subspace of the image space and that
distortions push them away from this subspace. BRISQUE
\cite{mittal2012brisque} extracts Generalized Gaussian Distribution
(GGD) features from normalized luminance coefficients to quantify blur
and noise. NIQE \cite{mittal2012niqe} constructs a Multivariate
Gaussian (MVG) model of natural image statistics and measures the
deviation of the test image from the natural-image manifold. These
methods are computationally efficient and mathematically interpretable,
but their hand-crafted features have limited capacity to capture
complex, unknown distortion manifolds, and they operate on the whole
image as a single quality unit.

\noindent\textbf{Deep CNN and transformer models.} The second
generation began with end-to-end quality regression. Bosse et al.
\cite{bosse2018deep} introduced a deep CNN with separate branches for
distortion and content, while Kang et al. \cite{kang2014cnniqa} proposed
one of the earliest CNN-based NR-IQA models. DBCNN \cite{zhang2020dbcnn}
combines a distortion branch and a content branch to capture both
low-level artifacts and high-level semantics. Transformer-based models
such as MUSIQ \cite{ke2021musiq} and MANIQA \cite{yang2022maniqa} leverage
multi-scale attention to capture long-range dependencies, while You and
Korhonen \cite{you2021transformer} adapt vision transformers directly to
NR-IQA. These models substantially improve accuracy on public benchmarks
such as CSIQ, LIVEC, and KonIQ-10k \cite{hosu2020koniq}, but they all
operate under the same monolithic assumption: the entire frame
constitutes a single quality unit.

\noindent\textbf{Vision-language IQA.} The third generation integrates
vision-language models into quality prediction. CLIP-IQA+
\cite{wang2023clip-iqa-plus} leverages CLIP's image-text alignment to
predict quality through textual prompts. Q-Align \cite{chen2024qalign}
teaches large multimodal models to score images via discrete
text-defined levels. TOPIQ \cite{chen2024topiq} adopts a top-down
approach from semantics to distortions. Zoom-IQA \cite{liang2026zoomiqa}
introduces reliable region-aware reasoning for quality assessment. More
recently, Liao et al. \cite{liao2026magnitude} observe that the magnitude
of CLIP features, not just the cosine similarity, is strongly correlated
with perceptual quality, and propose a magnitude-aware fusion scheme.
Ramesh et al. \cite{ramesh2025dgiqa} integrate scene depth into the
quality regression pipeline, showing improved generalization to natural
distortions such as haze and low-light. These models, however, also
operate on the whole image, without explicit target-level supervision.

Despite this progress, these methods share a common limitation: quality
is treated as a global property of the entire frame. When the target of
interest occupies only a small fraction of the image---as in railway
inspection, where the train body may cover less than 20\% of the
pixels---global aggregation is heavily dominated by irrelevant background
variables such as the sky, track ballast, or surrounding vegetation. This
monolithic assumption causes state-of-the-art whole-image models to
degrade severely in target-level evaluation---estimating the quality of
a specific target region rather than the whole image
(Section~\ref{sec:intro})---with their PLCC dropping to 0.8230 on our
benchmark (Section~\ref{sec:res-within}).

These three generations share a common whole-image assumption; the
following subsections review the lines that attempt to move beyond it.

\subsection{Multi-Teacher Aggregation and Consensus}
\label{sec:rw-consensus}

A common strategy for improving NR-IQA robustness is to aggregate the
outputs of multiple pretrained models. Multi-teacher distillation
transfers knowledge from an ensemble of teachers to a student network;
in the IQA context, a simpler alternative is consensus aggregation,
which computes a weighted average of teacher scores, often using
median-based or Gaussian-weighted aggregation to suppress outliers.
This strategy is attractive because it requires no additional training
and can leverage the complementary strengths of diverse IQA
backbones---NSS methods capture low-level statistics, CNN methods
capture learned distortion features, and CLIP-based methods capture
semantic quality. BiCG-Net follows the distillation route, but replaces
the heuristic average with in-image differencing
(Section~\ref{sec:observation}).

These methods, however, treat aggregation as a heuristic and assume
that individual teacher biases cancel out under averaging. Our
observation model (Section~\ref{sec:observation}) proves that this
assumption is flawed: the target term and the teacher bias are
entangled in each reading, and simple averaging inevitably propagates
both into the weak label. When the target occupies only a small
fraction of the frame, the bias and background terms dominate the
average, causing consensus to degrade; for example, under the
CSIQ-to-LIVEC shift, consensus achieves a PLCC of only 0.6469
(Section~\ref{sec:res-cross}), while our method reaches 0.8416. This
motivates in-image differencing (Section~\ref{sec:observation}), which
eliminates bias by construction rather than by statistical
cancellation.

\subsection{Region- and Target-Oriented Quality Assessment}
\label{sec:rw-region}

To address the spatial selectivity limitation of whole-image IQA, a line
of work has explored quality assessment on regions of interest (ROIs) or
specific semantic targets.

\noindent\textbf{Traffic scenarios.} RTSIQA \cite{lu2025rtsiqa} is the
most closely related work because it also targets region-level quality
in real-world scenes: it constructs a database and method specifically
for traffic scenes. However, RTSIQA still requires region-level MOS for
training and does not address cross-dataset transfer. Lu et al.
\cite{lu2025specific} evaluate quality in predefined areas of traffic
targets, but rely on static geometric priors.

\noindent\textbf{Patch-based methods.} Other methods use saliency maps,
manual bounding boxes, or patch decomposition to constrain the
evaluation region. PaQ-2-PiQ \cite{ying2020paq2piq} maps patch-level
quality to picture-level quality via learned aggregation. These methods
partially alleviate background contamination but still lack an explicit
mechanism for separating target quality from background contributions.
Patch-based aggregation typically computes a simple arithmetic mean
across patches, which propagates background interference from
non-target patches into the final score.

Existing region-oriented methods, therefore, still face three
challenges: (i) reliance on static or manually defined boundaries, (ii)
absence of an explicit background suppression mechanism, and (iii) lack
of cross-dataset transfer. Our method addresses all three through
in-image differencing (Section~\ref{sec:observation}) and adaptive
target localization (Section~\ref{sec:localization}).

\noindent\textbf{Localization and segmentation tools.} Precise target
localization and segmentation provide the spatial foundation for
target-specific quality assessment. YOLO \cite{redmon2016yolo}
introduced a unified object detection framework that localizes targets
with a single forward pass, and subsequent variants have made it a
standard tool for industrial inspection. The Segment Anything Model
(SAM) \cite{kirillov2023sam} enables prompt-driven, pixel-level
segmentation of arbitrary objects; combined with a bounding-box prompt
from an object detector, it produces high-quality masks that purify the
target region and remove background interference. In our framework,
YOLO and SAM are used only during offline preprocessing to localize
and purify the target, and the quality assessment itself does not
depend on the foundation models at inference time. This distinguishes
our approach from end-to-end foundation-model pipelines.

\subsection{Cross-Dataset Image Quality Assessment}
\label{sec:rw-cross}

The practical deployment of IQA systems requires generalization across
domains, yet most existing methods are evaluated under the assumption
that training and test data are drawn from the same distribution.

\noindent\textbf{Domain adaptation for IQA.} A number of works have
explored domain adaptation for IQA. MetaIQA \cite{zhu2020meta} applies
meta-learning to learn a prior that can be quickly adapted to new
domains. Madhusudana et al. \cite{madhusudana2022contrastive} leverage
contrastive learning to learn domain-invariant representations. These
methods typically require either target-domain data during adaptation or
a meta-training procedure, both of which are impractical in industrial
deployments where target-domain labels are unavailable.

\noindent\textbf{Cross-dataset distillation.} Gao et al.
\cite{gao2024cross} propose a cross-dataset distillation framework in
which a student aligns its representation with teachers trained on
different datasets, achieving an SRCC of 0.964 on CSIQ. However, it
requires access to both source and target domain data during training,
and relies on teacher models that are themselves trained on annotated
datasets.

\noindent\textbf{Cross-dataset transfer.} Other works study
cross-dataset IQA, where a model trained on one dataset is directly
applied to another. This corresponds to the L2 setting in our
experiments (Section~\ref{sec:res-cross}). These studies typically
report significant performance drops under domain shift, with PLCC
decreasing by 0.1--0.2 or more.

Existing cross-dataset IQA methods, however, either require
target-domain adaptation or report substantial performance degradation
under shift. In contrast, our method achieves cross-dataset transfer
without any target-domain labels, by exploiting a structural property of
teacher readings---the in-image differencing mechanism---that cancels
teacher bias within an image. We do not report zero-shot transfer from
the self-built dataset to the public benchmarks, because the two
settings differ in task definition: the self-built dataset is annotated
at the target level, whereas CSIQ and LIVEC provide whole-image MOS.
Such a transfer would conflate task mismatch with domain shift and is
therefore not directly comparable.

\subsection{Weak-Label and Pseudo-Label Generation}
\label{sec:rw-weak}

Beyond cross-dataset transfer, the supervision signal itself must be
generated without target-level MOS. This places our method in the
broader family of weakly supervised learning. Weak-label generation
produces supervision from model predictions, typically by selecting
high-confidence samples, and noisy label learning models label noise
explicitly to prevent corrupted supervision from degrading downstream
performance.

\noindent\textbf{Pseudo-labeling for IQA.} Recent work has explored
semi-supervised learning to reduce the annotation burden. CPL-IQA
\cite{zhong2025cpl} introduces a semi-supervised blind IQA (BIQA)
framework that combines a small set of labeled images with a large set
of unlabeled images. It converts MOS scalars into vector labels via
entropy minimization, then iteratively refines them using a
manifold-based label propagation strategy with confidence-quantifiable
pseudo-labels. This approach substantially reduces the need for manual
annotation while maintaining strong performance on authentic distortion
benchmarks.

CPL-IQA, however, assumes that a small set of labeled images is
available to initialize the label propagation process, and relies on
confidence estimation to filter unreliable pseudo-labels---a statistical
heuristic without a bias-elimination guarantee. In our target-specific
setting, no target-level human MOS exists. In contrast, in-image
differencing eliminates teacher bias by construction, without requiring
confidence estimation or labeled seed data.

\subsection{Positioning of This Work}
\label{sec:rw-positioning}

Having reviewed the related lines, we now position our work against
the most closely related ones. Table~\ref{tab:positioning} summarizes
the differences between our method and the most closely related lines
of work.

\begin{table}[htbp]
\centering
\small
\setlength{\tabcolsep}{4pt}
\caption{Comparison with related work.
Cross-data denotes cross-dataset transfer.}
\label{tab:positioning}
\begin{tabular}{lcccccc}
\toprule
Method & Target & No & Cross & Prov- & Multi- & Bg. \\
       & -level & MOS & -data & able & teacher & supp. \\
\midrule
Whole-image NR-IQA & $\times$ & $\times$ & $\times$ & $\times$ & $\times$ & $\times$ \\
Multi-teacher consensus & $\times$ & $\checkmark$ & $\times$ & $\times$ & $\checkmark$ & $\times$ \\
ROI-based IQA & $\checkmark$ & $\times$ & $\times$ & $\times$ & $\times$ & $\times$ \\
Cross-dataset IQA & $\times$ & $\times$ & $\checkmark$ & $\times$ & $\times$ & $\times$ \\
\textbf{BiCG-Net (Ours)} & $\checkmark$ & $\checkmark$ & $\checkmark$ & $\checkmark$ & $\checkmark$ & $\checkmark$ \\
\bottomrule
\end{tabular}
\end{table}

The key distinction is not only \emph{what} we estimate (target-level
quality without MOS) but \emph{how}: nested sampling turns background
contamination from a nuisance into a signal, which existing ROI-based
and whole-image methods do not exploit.

Despite the extensive literature reviewed above, three open problems
remain: (i) no MOS-free supervision for target-level IQA; (ii) no
provable bias elimination in multi-teacher aggregation; and (iii) no
bias-eliminated supervision for cross-dataset transfer.

To our knowledge, BiCG-Net is among the first frameworks that unify:
(i) target-level quality assessment, (ii) supervision without
target-level human MOS, (iii) cross-dataset transferability, (iv) a
provable bias-elimination mechanism, (v) multi-teacher aggregation, and
(vi) explicit background suppression. The in-image differencing property
introduced in Section~\ref{sec:method} provides the theoretical
foundation for (ii) and (iv), while the empirical results in
Section~\ref{sec:results} demonstrate (i), (iii), (v), and (vi).

\noindent\textbf{Comparison with RTSIQA.} RTSIQA \cite{lu2025rtsiqa}
constructs a database and method specifically for real-world traffic
scenes. While RTSIQA also targets region-level quality, it differs from
our method in three key aspects: (i) RTSIQA requires region-level MOS for
training, whereas our method requires no target-level MOS; (ii) RTSIQA
does not address cross-dataset transfer, whereas our method explicitly
evaluates it; and (iii) RTSIQA does not provide a formal
bias-elimination mechanism, whereas our in-image differencing provides
a provable foundation.
\section{Method}
\label{sec:method}

\subsection{Problem Formulation and Spatial Foundation}
\label{sec:problem}

\subsubsection{Target Localization and Nested Sampling}
\label{sec:localization}

Given an input image \(I \in \mathbb{R}^{W \times H \times 3}\), our
goal is to estimate the quality \(\hat Q\) of a \emph{specific target
region}, rather than that of the whole image. The target region is
specified by a bounding box \(\Bmin\) that covers the semantic object of
interest. To obtain \(\Bmin\), a pretrained YOLO detector
\cite{redmon2016yolo} first produces an initial box, which is then fed
as a prompt into the Segment Anything Model (SAM)
\cite{kirillov2023sam} to yield a pixel-level mask. The mask is cleaned
by morphological closing to fill small holes, and the minimum bounding
rectangle of the cleaned mask is taken as the refined box. The process
is iterated until the IoU between consecutive boxes falls below
\(\tau_{\text{IoU}}\). The final box \(\Bmin\) is treated as a fixed input
during quality assessment; both YOLO and SAM are applied offline and
are not fine-tuned on our data.

From \(\Bmin\), a set of \(\Npatch\) nested patches
\(\{B_k\}_{k=1}^{\Npatch}\) is generated by progressively expanding
outward, such that \(\Bmin \subseteq \Bk\) for all \(k\). For each patch, a
pretrained IQA teacher \(i\) (\(i = 1, \ldots, T\), with \(T = 11\))
produces a scalar reading \(f_i(\Bk)\). To describe the spatial
relationship between each patch and the target, we define three
geometric quantities:
\begin{align}
  \deltak &= 1 - \frac{|\Bmin|}{|\Bk|}, \quad \deltak \in [0, 1],
  \label{eq:delta} \\
  \text{area\_ratio}_k &= \frac{|\Bmin|}{|\Bk|} = 1 - \deltak,
  \label{eq:area-ratio} \\
  \text{offset}_k &= \sqrt{\left(\frac{cx_k - cx_{\min}}{W}\right)^2
                          + \left(\frac{cy_k - cy_{\min}}{H}\right)^2},
  \label{eq:offset}
\end{align}
where \((cx_k, cy_k)\) and \((cx_{\min}, cy_{\min})\) denote the centers of
\(\Bk\) and \(\Bmin\), respectively. These three quantities form the
geometric vector \(\gvec_k = [\deltak, \text{area\_ratio}_k,
\text{offset}_k] \in \mathbb{R}^3\). The first two enter the observation
model (Section~\ref{sec:observation}); all three are consumed by the
geometric encoder (Section~\ref{sec:correction}). Sampling patches are
sorted by \(\deltak\) in ascending order, so that the sequence preserves
the spatial cascading structure from the target outward.

\subsubsection{Notation}
\label{sec:notation}

Table~\ref{tab:notation} summarizes the symbols used throughout this
section.

\begin{table}[htbp]
\centering
\small
\caption{Notation.}
\label{tab:notation}
\setlength{\tabcolsep}{3pt}
\begin{tabular}{@{}ll@{\hspace{1.5em}}ll@{}}
\toprule
Symbol & Meaning & Symbol & Meaning \\
\midrule
\(I\) & Input image & \(s_i^{\text{eff}}\) & Effective per-teacher scale \\
\(B_{\min}\) & Target bounding box & \(\tilde{f}_i(B_k)\) & Normalized teacher reading \\
\(B_{\max}\) & Full-image box & \(\tilde{f}_i^{\text{corr}}(B_k)\) & Background-corrected reading \\
\(B_k\) & \(k\)-th sampling patch & \(\hat{\beta}_i\) & Predicted area coefficient \\
\(N_{\text{patch}}\) & Number of sampling patches & \(w_i^{\text{cred}}\) & Teacher credibility \\
\(T\) & Number of teachers (\(T=11\)) & \(d_k^{\text{dis}}\) & Cross-teacher dispersion \\
\(f_i(B_k)\) & Raw reading of teacher \(i\) & \(\mathbf{h}_k^{\text{geo}}\) & Geometric embedding \\
\(p_i\) & Polarity of teacher \(i\) (\(\pm1\)) & \(\mathbf{x}_k^{\text{seq}}\) & Sequence-encoder input \\
\(Q_{\text{target}}^*\) & Underlying target quality & \(\mathbf{z}^{\text{seq}}\) & Sequence feature \\
\(b_i\) & Systematic bias of teacher \(i\) & \(g_k^{\text{geo}}\) & Geometric gate \\
\(\beta_i\) & Area-effect slope & \(g_k^{\text{dis}}\) & Dispersion gate \\
\(\gamma_i\) & Background-content coefficient & \(g_k^{\text{agg}}\) & Combined aggregation gate \\
\(\delta_k\) & Difference-area ratio & \(\bar{f}_i\) & Aggregated teacher response \\
\(\phi_k\) & Background content & \(\hat{f}_i^{\text{tgt}}\) & Bias-free response \\
\(\epsilon_{i,k}\) & Measurement noise & \(\hat{Q}_{\text{raw}}\) & Raw quality head output \\
\(s_i^{\text{base}}\) & Global scale of teacher \(i\) & \(\lambda^{\text{qual}}, b^{\text{qual}}\) & Quality scale and bias \\
\(\eta_i\) & Learnable log-scale correction & \(\hat{Q}\) & Estimated target quality \\
\(\Delta\) & \(\hat{Q} - \text{consensus}\) & & \\
\bottomrule
\end{tabular}
\end{table}

\subsection{Observation Model and In-Image Differencing}
\label{sec:observation}

\noindent\textbf{Observation model.} We model the reading of teacher \(i\)
on sampling patch \(\Bk\) as a linear decomposition of five components:
\begin{equation}
f_i(\Bk) = p_i \Qtarget + b_i + \beta_i \deltak + \gamma_i \phik
         + \epsilon_{i,k}.
\label{eq:observation}
\end{equation}
The first term \(p_i \Qtarget\) captures the target quality contribution,
where \(p_i \in \{+1, -1\}\) encodes the polarity of the teacher. The
second term \(b_i\) is the systematic bias of teacher \(i\), which is
independent of the patch index \(k\). The third term \(\beta_i \deltak\)
accounts for the area effect, and the fourth term \(\gamma_i \phik\)
captures the background content. The fifth term \(\epsilon_{i,k}\) is
measurement noise.

\noindent\textbf{Underdetermined structure.} All five quantities in
Eq.~\eqref{eq:observation} are unobserved, and no direct separation is
possible even with pixel-perfect localization. This underdetermined
structure is the root cause of the failure of simple averaging across
teachers.

\noindent\textbf{In-image differencing.} Our core observation is that
although the five terms are individually unknown, two of them---the
target quality \(\Qtarget\) and the teacher bias \(b_i\)---are \emph{strictly
invariant within a single image}. In contrast, \(\deltak\) and \(\phik\)
vary dynamically with \(k\). Therefore, taking the difference between any
two patches \(B_j\) and \(B_k\) within the same image completely cancels
out the target term and the teacher bias.

\begin{proposition}[In-Image Differencing]
\label{prop:differencing}
For any two sampling patches \(B_j\) and \(B_k\) within the same image,
after subtracting an estimate of the area effect \(\beta_i \delta_k\)
(Section~\ref{sec:correction}), the following identity holds:
\begin{equation}
\fcorr_i(B_j) - \fcorr_i(B_k)
  = \gamma_i [\phij - \phik] + \epsilon_{i,j} - \epsilon_{i,k}.
\label{eq:differencing}
\end{equation}
The target term \(p_i \Qtarget\) and the systematic bias \(b_i\) are
exactly eliminated.
\end{proposition}

\begin{proof}
By Eq.~\eqref{eq:observation}, for patches \(B_j\) and \(B_k\):
\begin{align}
  f_i(B_j) - f_i(B_k)
  &= p_i \Qtarget + b_i + \beta_i \delta_j + \gamma_i \phij
     + \epsilon_{i,j} \nonumber \\
  &\quad - \left(p_i \Qtarget + b_i + \beta_i \delta_k
     + \gamma_i \phik + \epsilon_{i,k}\right) \nonumber \\
  &= \beta_i (\delta_j - \delta_k)
     + \gamma_i (\phij - \phik)
     + \epsilon_{i,j} - \epsilon_{i,k}.
\end{align}
After subtracting the area-effect term \(\hat\beta_i \deltak\) from each
reading, the residual difference becomes
Eq.~\eqref{eq:differencing}. Both \(p_i \Qtarget\) and \(b_i\) vanish
because they are identical for \(B_j\) and \(B_k\) within the same image.
\end{proof}

This property is experimentally validated in
Section~\ref{sec:res-cross}.

\noindent\textbf{Per-teacher scaling.} Before differencing, each
reading is polarity-corrected and normalized by a per-teacher scale
\(s_i^{\text{eff}}\), yielding
\(\tilde f_i(\Bk) = p_i f_i(\Bk) / s_i^{\text{eff}}\). Because
\(p_i / s_i^{\text{eff}}\) is a non-zero constant with fixed sign for
teacher \(i\), the ordering of the readings is preserved:

\begin{proposition}[Within-Teacher Order Preservation]
\label{prop:scale}
For any teacher \(i\) and patches \(B_j, B_k\),
\(\tilde f_i(B_j) > \tilde f_i(B_k) \iff f_i(B_j) > f_i(B_k)\).
\end{proposition}

\begin{proof}
\(\tilde f_i(B_j) - \tilde f_i(B_k) = (p_i / s_i^{\text{eff}})
(f_i(B_j) - f_i(B_k))\), and \(p_i / s_i^{\text{eff}}\) has fixed sign
within teacher \(i\).
\end{proof}

\noindent\textbf{Within-image invariance.} Because the differencing is
performed \emph{within} an image, both the target quality term
\(p_i \Qtarget\) and the teacher bias \(b_i\) are eliminated from the
residual. The resulting statistics therefore depend only on the
\emph{relative} background content across patches, not on the absolute
scale or bias of the teacher. This within-image invariance is a
necessary condition for cross-benchmark transfer without target-domain
labels; the empirical evidence is provided in
Section~\ref{sec:res-cross}.

\noindent\textbf{Consensus baseline.} Throughout the paper, we use the
arithmetic mean of the \(T\) teacher readings on the target box as the
consensus baseline:
\begin{equation}
\mathrm{consensus} = \frac{1}{T} \sum_{i=1}^{T} f_i(\Bmin).
\label{eq:consensus}
\end{equation}
The consensus represents the knowledge that a naive multi-teacher
aggregation would transfer. For weak-label generation on the self-built
dataset, we use a Gaussian-weighted variant, described in
Section~\ref{sec:loss}.

\subsection{Network Architecture}
\label{sec:architecture}

\begin{figure}[htbp]
\centering
\includegraphics[width=0.95\textwidth]{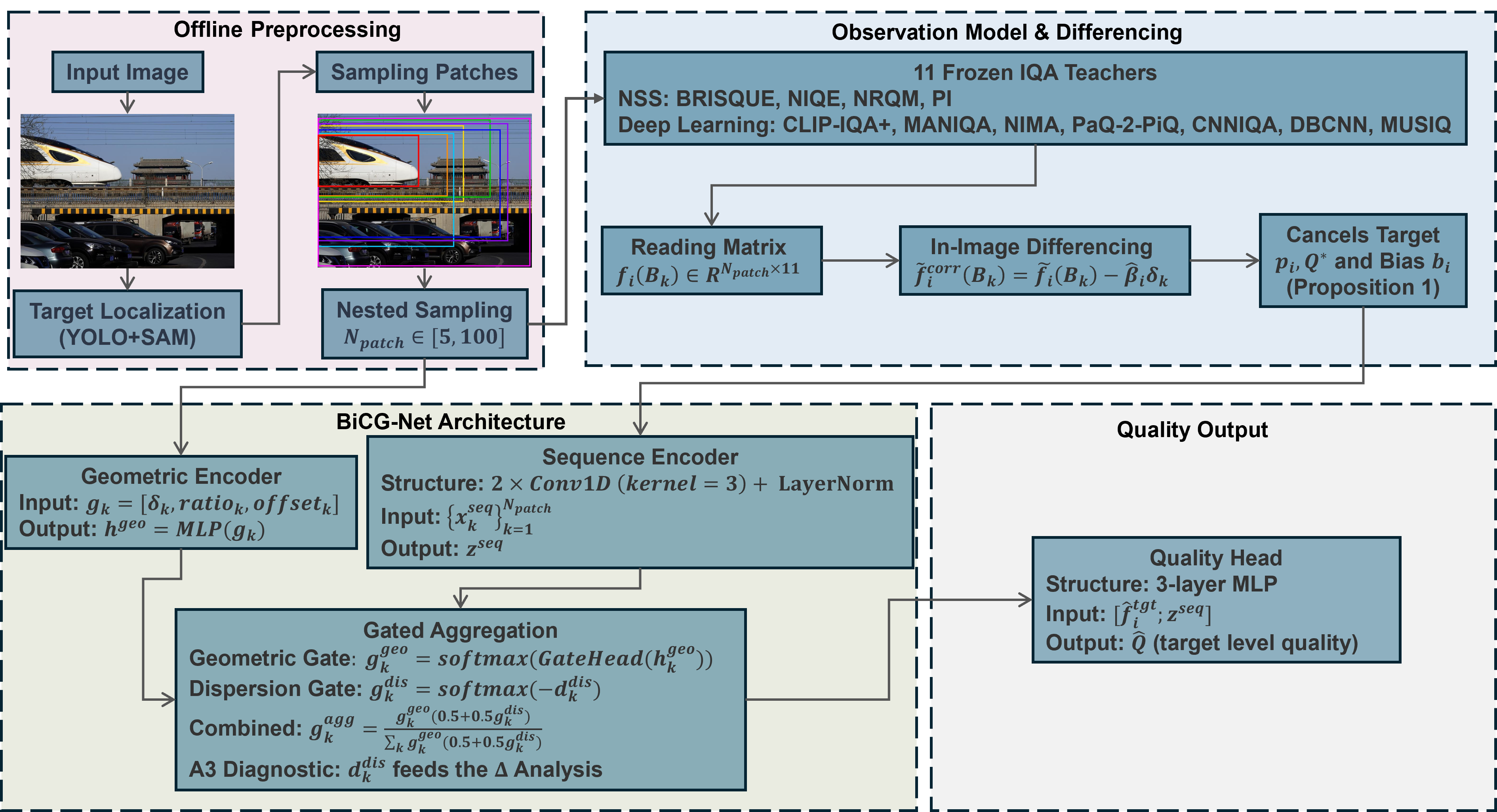}
\caption{Overview of the BiCG-Net pipeline. The target box is localized
by YOLO and SAM, expanded into a nested patch sequence sorted by
\(\delta_k\), read by 11 frozen IQA teachers, corrected, encoded, gated,
and mapped to the target-level score \(\hat Q\).}
\label{fig:pipeline}
\end{figure}

\noindent\textbf{Overview.} Figure~\ref{fig:pipeline} summarizes the
pipeline, which consists of four modules operating on the intra-image
statistics derived from Section~\ref{sec:observation}. The first three
modules correspond to specific terms of the observation model:
per-teacher scale (Section~\ref{sec:scale}), area-effect correction
(Section~\ref{sec:correction}), and sequence encoding
(Section~\ref{sec:encoder}). The fourth module
(Section~\ref{sec:aggregation}) aggregates the corrected readings
through two gates. The dispersion gate in this final module is
deliberately designed as a diagnostic pathway (feeding the \(\Delta\)
analysis in Section~\ref{sec:delta}) rather than an accuracy-oriented
one; its accuracy contribution is quantified separately in the ablation
study of Section~\ref{sec:res-ablation}.

\subsubsection{Per-Teacher Scale}
\label{sec:scale}

The readings of different teachers are on different numerical scales.
We first apply the polarity \(p_i\) offline, so that all teacher readings
follow the convention ``higher is better.'' In the network, the
polarity-corrected reading is normalized by a global scale estimated
from the training set:
\begin{equation}
\sbase_i = \text{median}_{g, k} \left| f_i(\Bk) \right|.
\end{equation}
A learnable log-scale correction \(\eta_i\) is then applied:
\begin{equation}
\seff_i = \sbase_i \cdot \exp(\eta_i),
\label{eq:scale-eff}
\end{equation}
where \(\eta_i\) is initialized to zero and \(\seff_i\) is clamped to
\([10^{-6}, 10^{6}]\) to prevent numerical instability. The normalized
reading is
\begin{equation}
\ftilde_i(\Bk) = \frac{p_i \cdot f_i(\Bk)}{\seff_i}.
\label{eq:scale-normalize}
\end{equation}

\subsubsection{Observation Correction}
\label{sec:correction}

The geometric vector \(\gvec_k\) is encoded into a 64-dimensional
embedding through a two-layer MLP:
\begin{equation}
\hgeo_k = \text{GeoEncoder}(\gvec_k) \in \mathbb{R}^{64}.
\end{equation}
To maintain a stable and robust background correction across all
patches, a per-image area coefficient is predicted by pooling the
geometric embeddings:
\begin{equation}
\hat\beta_i = \text{TeacherBGEncoder}\left(\frac{1}{\Npatch}\sum_{k=1}^{\Npatch} \hgeo_k\right).
\label{eq:beta}
\end{equation}
The background correction is then applied:
\begin{equation}
\fcorr_i(\Bk) = \ftilde_i(\Bk) - \hat\beta_i \cdot \deltak.
\label{eq:correction}
\end{equation}
This correction removes the area-induced bias from the raw reading,
leaving only the target quality, systematic bias, background content,
and noise.

The teacher credibility is computed from the within-image variance of
the corrected readings:
\begin{equation}
\wcred_i = \mathrm{softmax}\left(-\mathrm{Var}_k[\fcorr_i(\Bk)]\right).
\label{eq:credibility}
\end{equation}

\subsubsection{Sequence Encoder}
\label{sec:encoder}

Because sampling patches form a nested sequence, their local patterns
contain structural information that simple aggregation cannot capture.
Each patch is encoded into a \((T + 2)\)-dimensional vector:
\begin{equation}
\xseq_k = \left[\fcorr_1(\Bk), \ldots, \fcorr_T(\Bk), \deltak, \ddis_k\right]
      \in \mathbb{R}^{T + 2},
\label{eq:seq-input}
\end{equation}
where
\begin{equation}
\ddis_k = \text{Var}_i \left[\fcorr_i(\Bk)\right]
\end{equation}
is the cross-teacher dispersion on patch \(\Bk\). The sequence
\(\{\xseq_k\}_{k=1}^{\Npatch}\) is processed by two 1D convolutional layers
with kernel size 3, padding 1, and 64 output channels each:
\begin{align}
  \Hconv{1} &= \text{ReLU}(\text{Conv1D}_{64}(\mathbf{X})), \\
  \Hconv{2} &= \text{ReLU}(\text{Conv1D}_{64}(\Hconv{1})),
\end{align}
followed by a masked mean pooling and a layer normalization to yield
the final sequence feature:
\begin{equation}
\zseq = \text{LayerNorm}
  \left(\frac{\sum_k \text{mask}_k \cdot \Hconv{2}_k}
              {\sum_k \text{mask}_k}\right) \in \mathbb{R}^{64}.
\label{eq:seq-feat}
\end{equation}

\subsubsection{Target Quality Aggregation}
\label{sec:aggregation}

To aggregate the corrected readings, we introduce two adaptive gates.
First, a geometric gate emphasizes patches whose spatial layout is most
representative of the core target:
\begin{equation}
g_k^{\text{geo}} = \text{softmax}_k\left(\text{GateHead}(\hgeo_k)\right),
\label{eq:gate-geo}
\end{equation}
where the mask is applied before the softmax to ensure padded patches
receive zero weight. Second, a dispersion gate acts as a reliability
filter, downweighting patches with high cross-teacher disagreement:
\begin{equation}
g_k^{\text{dis}} = \text{softmax}_k(-\ddis_k), \quad
\tilde g_k^{\text{dis}} = 0.5 + 0.5 \cdot g_k^{\text{dis}} \in [0.5, 1.0].
\label{eq:gate-dis}
\end{equation}
We map the dispersion gate to \([0.5, 1.0]\) to prevent complete
suppression, as occasional teacher disagreements may occur even for
high-quality targets due to differing feature biases. The two gates are
combined and renormalized:
\begin{equation}
\gagg_k = \frac{g_k^{\text{geo}} \cdot \tilde g_k^{\text{dis}}}
           {\sum_{k'} g_{k'}^{\text{geo}} \cdot \tilde g_{k'}^{\text{dis}}}.
\label{eq:gate-combine}
\end{equation}
The aggregated teacher response is
\begin{equation}
\fagg_i = \sum_k \gagg_k \cdot \fcorr_i(\Bk).
\label{eq:aggregate}
\end{equation}
The teacher bias \(b_i\) is predicted by a two-layer MLP bias head from
the target-box readings:
\begin{equation}
b_i = \text{BiasHead}(\ftilde_i(\Bmin)).
\end{equation}
The bias-free aggregated response and the raw quality estimate are:
\begin{equation}
\ftgt_i = \fagg_i - b_i,
\end{equation}
\begin{equation}
\hat Q_{\text{raw}} = \text{QualityHead}
  \left([\ftgt_i; \zseq]\right),
\label{eq:q-raw}
\end{equation}
where QualityHead is a three-layer MLP with hidden dimension 256.
The final target quality estimate is
\begin{equation}
\hat Q = \lqual \cdot \hat Q_{\text{raw}} + \bqual,
\label{eq:q-hat}
\end{equation}
where \(\lqual\) and \(\bqual\) are learnable scalar parameters initialized
to 1 and 0, respectively.

\subsection{Loss Function and Kendall log1p}
\label{sec:loss}

The network is trained with six complementary losses that jointly
constrain the corrected readings, the aggregated response, and the
final quality estimate:
\begin{align}
  L_{\text{bg}} &= \text{mean}_i \text{Var}_k \left[\fcorr_i(\Bk)\right], \\
  L_{\text{cross}} &= \text{mean}_k \text{Var}_i \left[\fcorr_i(\Bk)\right], \\
  L_{\text{target}} &= \text{MSE}(\hat Q, \mathrm{weak\_label}), \\
  L_{\text{bias}} &= \text{mean}_i b_i^2, \\
  L_{\text{coupling}} &= \text{MSE}(\fagg_i, \hat Q_{\text{raw}} + b_i), \\
  L_{\text{scale}} &= \text{mean}_i \eta_i^2.
\end{align}
The \(L_{\text{target}}\) term anchors the prediction to the absolute
quality scale, while the background structure is captured by the
remaining five losses and the gating module.

\noindent\textbf{Supervision signal.} For CSIQ and LIVEC, the human MOS
is used directly as \(\mathrm{weak\_label}\). For the self-built dataset,
since no target-level MOS is available, the weak label is generated by
the procedure below.

\noindent\textbf{Weak-label generation.} For the self-built dataset,
we first fit a within-image linear model
\(\ftilde_i(\Bk) = \alpha_i + \beta_i \deltak\) for each teacher and
subtract the area effect to obtain \(\hat{f}_i(\Bk)\). The teachers are
partitioned into two groups: \(F_B\) (NSS-based and hand-crafted) and
\(F_C\) (deep-learning). A Gaussian-weighted consensus is computed within
each group, denoted \(S_b\) and \(S_c\) respectively. The balanced weak
label is
\begin{equation}
\mathrm{weak\_label} = 0.5 \cdot S_b + 0.5 \cdot S_c,
\label{eq:weak-label}
\end{equation}
where the \(0.5/0.5\) weighting prevents either teacher family from
dominating the supervision signal.

\noindent\textbf{Kendall log1p correction.} To balance these losses
automatically, we adopt the Kendall homoscedastic uncertainty framework
\cite{kendall2018multi}:
\begin{equation}
L = \sum_{j \in \mathcal{J}} \left(\frac{L_j}{2\sigma_j^2} + \log \sigma_j\right),
\label{eq:kendall-original}
\end{equation}
where \(\mathcal{J} = \{\text{bg}, \text{cross}, \text{target},
\text{bias}, \text{coupling}, \text{scale}\}\). The original formulation
is unbounded below: as \(\sigma_j \to 0\), the term \(\log \sigma_j \to -\infty\).
We replace the logarithmic penalty with a log1p form:
\begin{equation}
L = \sum_{j \in \mathcal{J}} \left(\frac{L_j}{2\sigma_j^2}
                 + \alpha \cdot \log(1 + \sigma_j^2)\right),
\quad \alpha = 0.5.
\label{eq:kendall-log1p}
\end{equation}

\begin{proposition}[Boundedness of Kendall log1p]
\label{prop:kendall}
For each loss term \(j\) with \(L_j \ge 0\), the function
\(g_j(\sigma_j^2) = L_j/(2\sigma_j^2) + \alpha\log(1+\sigma_j^2)\)
(with \(\alpha > 0\)) admits a unique global minimum at
\(\sigma_j^{2*} > 0\) satisfying \(\alpha (\sigma_j^{2*})^2
= L_j (1+\sigma_j^{2*})/2\). Consequently, the total loss
Eq.~\eqref{eq:kendall-log1p} is bounded below.
\end{proposition}

\begin{proof}
Let \(u = \sigma_j^2 > 0\). Then
\(g_j(u) = L_j/(2u) + \alpha\log(1+u)\).
The derivative is
\(g_j'(u) = -L_j/(2u^2) + \alpha/(1+u)\).
Setting \(g_j'(u) = 0\) yields
\(\alpha u^2 = L_j(1+u)/2\), which has a unique positive root
\(u^*\) (the left-hand side is increasing in \(u\) and the right-hand
side is linear). Since \(g_j'(u) < 0\) for \(u < u^*\) and \(g_j'(u) > 0\)
for \(u > u^*\), \(u^*\) is the unique global minimum. Hence \(g_j\) is
bounded below, and so is the total loss.
\end{proof}

\noindent\textbf{Optimization.} We use AdamW with \(10^{-3}\) for model
parameters and \(10^{-2}\) for \(\sigma_j\), weight decay \(10^{-4}\), and
gradient clipping at 1.0. The \(\sigma_j\) are initialized to zero.

\subsection{\texorpdfstring{\(\Delta\)}{Delta} Structure Analysis}
\label{sec:delta}

The training losses do not directly reveal what the student has learned
beyond the teachers. To diagnose this, we define the residual
\(\Delta = \hat Q - \mathrm{consensus}\). Assuming both estimators
approximate \(\Qtarget\) with independent errors \(\etastu\) and
\(\etacons\), the residual becomes \(\Delta = \etastu - \etacons\).

\begin{proposition}[Elimination of \(\Qtarget\)]
\label{prop:delta}
The residual \(\Delta\) does not contain the target quality term \(\Qtarget\).
\end{proposition}

\begin{proof}
Let \(\hat Q = \Qtarget + \eta^{\text{stu}}\) and
\(\mathrm{consensus} = \Qtarget + \eta^{\text{cons}}\), where
\(\eta^{\text{stu}}\) and \(\eta^{\text{cons}}\) denote the estimation
errors. Substituting into \(\Delta = \hat Q - \mathrm{consensus}\) yields
\(\Delta = \eta^{\text{stu}} - \eta^{\text{cons}}\), which is independent
of \(\Qtarget\).
\end{proof}

Any significant correlation between \(\Delta\) and a physical quantity
indicates that BiCG-Net and consensus differ along that dimension.
Let \(B_{\max}\) denote the full-image box. We consider five physically
interpretable quantities:
\begin{align}
\Dtgtbg &= \sum_i \omega_i [\fcorr_i(\Bmin) - \fcorr_i(B_{\max})],
  \label{eq:V1} \\
\Dclean &= \text{mean}_k \text{Var}_i[\fcorr_i(\Bk)]
       - \text{mean}_i[\epsilon_{i,k}^2], \\
\Cintra &= 1 - \frac{\text{Var}_k [\fcorr_i(\Bk)]}{\text{mean}_k [\fcorr_i(\Bk)]}, \\
\Ewl    &= |\mathrm{weak\_label} - 0.5|, \\
\Sarea  &= \text{mean}_i |\hat\beta_i|.
\end{align}
We exclude three model-internal quantities (teacher-credibility
entropy, sampling count, and geometric-gate entropy) because they are
not physical quantities and are highly correlated with \(\Dtgtbg\) and
\(\Dclean\) in preliminary runs.

Significant correlations between \(\Delta\) and any of the retained
quantities provide direct evidence that BiCG-Net extracts information
beyond consensus. The \(\Delta\) analysis is reported on CSIQ and LIVEC,
the two benchmarks with human MOS; on the self-built dataset no
target-level MOS exists, so \(\Delta\) is interpreted against geometric
and dispersion quantities in Section~\ref{sec:res-within}. The
experimental results are reported in Section~\ref{sec:ana-delta}.
\section{Experiments}
\label{sec:experiments}

\subsection{Datasets}
\label{sec:datasets}

We evaluate on three datasets: one self-built railway dataset and two
public benchmarks (CSIQ and LIVEC). Table~\ref{tab:task-def}
summarizes the task definitions. \textbf{Each dataset is trained and
tested independently (L1); no model trained on one dataset is
evaluated on another, except in the explicit cross-dataset transfer
experiments of Section~\ref{sec:res-cross} (L2), where the
source-domain student is frozen and applied zero-shot.}

\begin{table}[htbp]
\centering
\small
\caption{Task definition across the three datasets. The self-built
dataset is a genuine target-level task; CSIQ and LIVEC are evaluated
under the degenerated whole-image setting. All within-domain (L1)
results train and test on the same dataset; cross-dataset (L2) results
are reported separately in Table~\ref{tab:res-cross}.}
\label{tab:task-def}
\begin{tabular}{lccc}
\toprule
 & Self-built & CSIQ & LIVEC \\
\midrule
Train/Test & Same dataset & Same dataset & Same dataset \\
Input & target box + patches & whole image & whole image \\
Output & target-level quality & frame-level MOS & frame-level MOS \\
Supervision & weak label & human MOS & human MOS \\
Split & image-level & reference-level & random \\
\bottomrule
\end{tabular}
\end{table}

\noindent\textbf{Self-built dataset.} This dataset comprises 2,816 high-resolution images captured from unconstrained railway scenes. Each image is annotated with a target bounding box enclosing the train body, which is obtained via YOLO + SAM + IoU refinement (Section~\ref{sec:localization}). The objective of this task is to estimate target-level quality. Given that no target-level human MOS is available, the supervision signal relies on a weak label generated by the consensus algorithm (Section~\ref{sec:loss}). The dataset is split at the image level into 2,252 training images and 564 test images using an 8:2 ratio with seed 42. Out of the 2,252 training images, 2,235 are valid, while 17 are skipped because their target boxes yield fewer than the minimum number of sampling patches. Similarly, among the 564 test images, 559 are valid, and the remaining 5 are skipped for the same reason. All reported test metrics on this self-built dataset are computed directly against the generated weak labels.

The two tracks serve complementary roles: CSIQ and LIVEC provide
perceptual ground truth for validating the differencing mechanism
against human MOS, while the self-built dataset validates the same
mechanism against a weak-label consensus in the deployment scenario.
The shared architecture and differencing mechanism link the two: the
MOS benchmarks establish perceptual validity, and the self-built
results demonstrate that this validity transfers to the unannotated
industrial setting.

\noindent\textbf{CSIQ.} 30 reference images, each distorted by six
distortion types (blur, contrast, fnoise, jpeg, jpeg2000, noise) at
five levels, yielding 900 images. After removing 34 images without
MOS, 866 valid images remain.\footnote{The 34 removed images
correspond to the contrast distortion at levels 4 and 5, for which no
MOS is available in the original CSIQ annotation.} The dataset is
split at the reference-image level into 24 training references (693
images) and 6 test references (173 images). Task: \emph{whole-image
MOS}.

\noindent\textbf{LIVEC.} 1,162 authentic images with realistic
distortions, no reference images, no distortion labels. The dataset is
split randomly at the image level into 929 training images and 233
test images (8:2). Task: \emph{whole-image MOS}.

\noindent\textbf{Why these three, and how they are split.} The
self-built dataset represents the deployment scenario (target-level
quality); CSIQ provides a controlled distortion manifold; LIVEC
provides an unconstrained benchmark. Together they cover the spectrum
from controlled to in-the-wild settings. Because the three dataset
structures differ, their splitting strategies also differ; all splits
use an 8:2 ratio and the same random seed (42) to ensure
reproducibility.

\noindent\textbf{Target boxes for the public datasets.} For CSIQ and
LIVEC, no target-level MOS is available. We therefore use a centered
box with an area ratio of $1/9$ as the target box $\Bmin$, so that the
target-level task degenerates to whole-image quality assessment.

\subsection{Implementation Details}
\label{sec:implementation}

\noindent\textbf{Teacher ensemble.} The 11 teachers cover three
complementary categories: classical NSS-based (BRISQUE
\cite{mittal2012brisque}, NIQE \cite{mittal2012niqe}), hand-crafted
perceptual (NRQM \cite{ma2017nrqm}, PI \cite{blau2018perception}), and
deep learning (CLIP-IQA+ \cite{wang2023clip-iqa-plus}, MANIQA
\cite{yang2022maniqa}, NIMA \cite{talebi2018nima}, PaQ-2-PiQ
\cite{ying2020paq2piq}, CNNIQA \cite{kang2014cnniqa}, DBCNN
\cite{zhang2020dbcnn}, MUSIQ \cite{ke2021musiq}). No teacher is
fine-tuned, so the student learns from a fixed ensemble. The polarity
$p_i$ of each teacher is fixed offline: BRISQUE, NIQE, NRQM, and PI are
negative (lower is better), while the remaining seven are positive.

\noindent\textbf{Student network.} BiCG-Net instantiates the four
modules of Section~\ref{sec:architecture} with a compact design
(39,482 trainable parameters), orders of magnitude smaller than the
teacher ensemble it distills from. The geometric encoder and teacher
background encoder are two-layer MLPs with 64 and 128 hidden units; the
sequence encoder has two 1D convolutional layers with 64 channels and
kernel size 3; the quality head is a three-layer MLP with hidden
dimension 256.

\noindent\textbf{Data processing.} For each image, the target box is
first localized by YOLOv8 and refined by SAM~2.0
(Section~\ref{sec:localization}); both models are applied offline and
are not fine-tuned on our data. Then $\Npatch \in [10, 100]$ sampling
patches are generated by expanding outward from the target box and
sorted by $\deltak$ in ascending order. Each teacher is applied to
every patch, yielding a reading matrix of size $\Npatch \times T$. The
readings are polarity-corrected, normalized by the per-teacher scale,
and corrected for the area effect (Section~\ref{sec:correction}). In
the extreme off-center analysis of Section~\ref{sec:design}, the
nested subset is deliberately pushed down to $\Npatch \in [5, 7]$.

\noindent\textbf{Training.} AdamW is used with $10^{-3}$ for model
parameters and $10^{-2}$ for log-sigma parameters, weight decay
$10^{-4}$, and gradient clipping at 1.0. Epochs are set to 150 for the
self-built dataset and 100 for CSIQ/LIVEC; the best checkpoint is
selected by test PLCC, which introduces a mild optimistic bias in the
reported within-domain numbers. No data augmentation is applied.

\begin{table}[htbp]
\centering
\small
\caption{Training configuration.}
\label{tab:hyperparams}
\begin{tabular}{lccc}
\toprule
Parameter & Self-built & CSIQ & LIVEC \\
\midrule
Epochs & 150 & 100 & 100 \\
Best epoch & 110 & 50 & 100 \\
Batch size & 8 & 8 & 8 \\
Learning rate & $10^{-3}$ & $10^{-3}$ & $10^{-3}$ \\
$\sigma$ learning rate & $10^{-2}$ & $10^{-2}$ & $10^{-2}$ \\
Weight decay & $10^{-4}$ & $10^{-4}$ & $10^{-4}$ \\
$\alpha$ (Kendall) & 0.5 & 0.5 & 0.5 \\
Gradient clip & 1.0 & 1.0 & 1.0 \\
Sequence dim & 64 & 64 & 64 \\
Hidden dim & 256 & 256 & 256 \\
Eval interval & 5 & 5 & 5 \\
Supervision & Weak label & MOS & MOS \\
Split & Image-level & Reference-level & Random \\
\bottomrule
\end{tabular}
\end{table}

\noindent\textbf{Computational cost.} The teacher ensemble dominates
the offline pipeline, accounting for 99.96\% of the total latency
(Table~\ref{tab:cost-breakdown}); this cost is incurred only once during
dataset construction. The student network's inference overhead is
negligible, and the pipeline bottleneck resides in the teacher
ensemble. All timings are measured on dual NVIDIA GTX 1080 Ti GPUs with
32~GB RAM and an Intel Core i7-6800K CPU.

\subsection{Experimental Design and Rationale}
\label{sec:design}

The design answers three questions: (i) is the student network
effective \emph{within} a domain? (ii) does it generalize \emph{across}
datasets? (iii) which modules are \emph{necessary}? Table~\ref{tab:exp-design}
summarizes the three lines.

\begin{table}[htbp]
\centering
\small
\caption{Three experimental lines.}
\label{tab:exp-design}
\begin{tabular}{llll}
\toprule
Line & Train & Test & Tests \\
\midrule
L1 & Same dataset & Same dataset & Within-domain effectiveness \\
L2 & CSIQ/LIVEC & LIVEC/CSIQ & Cross-dataset transfer \\
L3 & Same dataset & Same dataset & Module necessity \\
\bottomrule
\end{tabular}
\end{table}

\noindent\textbf{Baselines.} We compare against three types of
baselines:
\begin{itemize}
  \item \textbf{Best single teacher.} The strongest individual IQA
        model on each dataset. All teacher models are evaluated using
        their \emph{official pre-trained weights} without any
        fine-tuning on our self-built railway dataset. On CSIQ and
        LIVEC, these weights correspond to models that may have been
        exposed to similar data during their original training,
        representing a strong, domain-adapted upper bound.
  \item \textbf{Teacher consensus.} The arithmetic mean of the 11
        teacher readings (Eq.~\eqref{eq:consensus}), representing the
        knowledge that a naive multi-teacher aggregation would
        transfer.
  \item \textbf{State-of-the-art whole-image NR-IQA models.} Models
        such as MUSIQ and MANIQA, testing whether our framework
        exceeds the direct output of individual foundation models.
\end{itemize}

\noindent\textbf{L1: Within-domain.} Train and test on the same
dataset, establishing the upper-bound performance when training and
test distributions match.

\noindent\textbf{L2: Cross-dataset transfer.} Train on one public
dataset, test on the other, without any target-domain fine-tuning or
labels. Both datasets have whole-image MOS, so the comparison is not
confounded by supervision quality. In L2, the student is evaluated
under a strict zero-shot setting: the model weights, per-teacher base
scales, and log-sigma parameters are all frozen at their source-domain
values.

\noindent\textbf{Baseline asymmetry in L2.} The teacher models are
evaluated with their official pre-trained weights, which likely overlap
with the target datasets, giving them an inherent advantage. In
contrast, our student network is strictly zero-shot: it is applied to
the target domain without any exposure to its data. Therefore, any
cross-dataset gains over the teachers indicate that the in-image
differencing mechanism cancels teacher bias under the CSIQ-to-LIVEC
shift. We do not report zero-shot transfer from the self-built dataset
to the public benchmarks, because the two settings differ in task
definition (target-level versus whole-image MOS); such a transfer would
conflate task mismatch with domain shift and is therefore not directly
comparable.

\noindent\textbf{L3: Ablation.} Disable each module in turn and
measure the performance drop across all three datasets to verify that
the conclusions are not dataset-specific. The full configuration list
(A0--A7) is reported in Section~\ref{sec:res-ablation}.

\noindent\textbf{Off-center anchor design.} Due to spatial jitter in
the fixed 50-box template, strict containment breaks. We adopt
domain-specific stratification: CSIQ uses percentile anchors
(25/50/75\%) pushing the offset to 0.115; LIVEC uses percentile
extraction under a sparse manifold, yielding offsets of 0.052, 0.080,
and 0.107. Both push $\Npatch \in [5,7]$, probing the boundary where
spatial shift and information scarcity coincide.
\section{Results}
\label{sec:results}

\subsection{Within-Domain Results and Main Findings}
\label{sec:res-within}

We first evaluate the within-domain setting (L1), where training and
test distributions match. Table~\ref{tab:main-results} reports the main
results across the three datasets. All numbers are computed from the
JSON/CSV outputs of the training and evaluation scripts; no post-hoc
adjustments are applied.

\begin{table}[htbp]
\centering
\small
\caption{Main results on the three datasets.}
\label{tab:main-results}
\begin{tabular}{lcccccc}
\toprule
 & \multicolumn{2}{c}{Self-built} & \multicolumn{2}{c}{CSIQ} & \multicolumn{2}{c}{LIVEC} \\
\cmidrule(lr){2-3} \cmidrule(lr){4-5} \cmidrule(lr){6-7}
Method & PLCC & MAE & PLCC & MAE & PLCC & MAE \\
\midrule
Best single teacher & 0.8230 & 1.2989 & 0.7715 & 0.1354 & 0.8443 & 0.0948 \\
Teacher consensus & 0.8642 & 0.0551 & 0.7427 & 0.1385 & 0.6469 & 0.1375 \\
\textbf{BiCG-Net} & \textbf{0.9855} & \textbf{0.0198} & \textbf{0.9009} & \textbf{0.0891} & \textbf{0.9133} & \textbf{0.0719} \\
\bottomrule
\end{tabular}

\vspace{1mm} 
\begin{flushleft}
\footnotesize \textit{Note:} The self-built dataset is evaluated against the weak label, whereas CSIQ and LIVEC are evaluated against human MOS. Since the weak label is derived from a Gaussian-weighted consensus, the consensus baseline naturally has a low MAE on the self-built dataset; its lower PLCC indicates a poorer rank ordering.
\end{flushleft}
\end{table}

\noindent\textbf{Why the student exceeds the teachers.} On all three
datasets, the student outperforms both the best single teacher and the
teacher consensus. This may seem counter-intuitive for a distilled
model, but it follows directly from the training objective: the student
is not trained to mimic any individual teacher, but to predict the
ground-truth label from the \emph{relative structure} of the teacher
readings. The $L_{\text{target}}$ term anchors the prediction to the
absolute quality scale, while the background structure is captured by
the remaining five losses. By aggregating complementary information
across teachers and patches, the student extracts intra-image cues that
no single teacher or consensus average can represent.

\noindent\textbf{On the self-built weak label.} A naive distillation
student fitted to a weak label derived from the teacher consensus would
trivially achieve a high PLCC, which could be mistaken for a circular
validation. However, our architecture and experimental setup explicitly
break this circularity. First, the student is not trained to minimize
the distance to individual teacher outputs; it minimizes \(L_{\text{target}}\)
against the consensus, while the remaining five losses enforce
structural invariance and background suppression (Section~\ref{sec:loss}).
Second, the achieved PLCC of 0.9855 is not merely a fit to the
consensus: the $\Delta$ analysis (Table~\ref{tab:delta-self}) shows that
the residual \(\Delta = \hat{Q} - \mathrm{weak\_label}\) is
significantly correlated with physically interpretable quantities
(e.g., $\Dtgtbg$ and $\Dclean$). If the student were simply
memorizing the consensus, this residual would be pure noise and
uncorrelated with any physical quantity. Therefore, the high PLCC on
the self-built dataset demonstrates the model's ability to capture
the underlying structure of the weak supervision, not a circular
reproduction of the teacher outputs.

\noindent\textbf{Self-built $\Delta$ analysis.} Following the same
diagnostic logic as in Section~\ref{sec:ana-delta}, we analyze the
residual on the self-built dataset. Since no target-level MOS exists,
$\Delta = \hat Q - \mathrm{weak\_label}$ is interpreted against
geometric and dispersion quantities. Table~\ref{tab:delta-self} reports
the Pearson correlation between $\Delta$ and five quantities on the 559
valid test images. The residual is significantly correlated with
$\Dtgtbg$ ($r=-0.3921$, $p<0.001$), $\Dclean$ ($r=+0.5526$,
$p<0.001$), and teacher-credibility entropy ($r=+0.1126$, $p=0.008$),
while the correlation with $\Cintra$ ($p=0.237$) and $\Sarea$
($p=0.772$) is not significant. This pattern indicates that on the
self-built dataset the student departs from the weak label mainly along
the background-difference and teacher-dispersion axes, consistent with
the MOS-benchmark findings in Section~\ref{sec:ana-delta}.

\begin{table}[htbp]
\centering
\small
\caption{Self-built \(\Delta\) analysis (559 test images), where \(\Delta = \hat{Q} - \mathrm{weak\_label}\).}
\label{tab:delta-self}
\begin{tabular}{lcc}
\toprule
Quantity & \(r\) (with \(\Delta\)) & \(p\) \\
\midrule
\(\Dtgtbg\) & \(-0.3921\) & \(<0.001\) \\
\(\Dclean\) & \(+0.5526\) & \(<0.001\) \\
\(H_w\) & \(+0.1126\) & \(0.008\) \\
\(\Cintra\) & \(-0.0467\) & \(0.237\) \\
\(\Sarea\) & \(+0.0083\) & \(0.772\) \\
\bottomrule
\end{tabular}

\vspace{2mm}
\begin{flushleft}
\footnotesize \textit{Note:} \(\Dtgtbg\): target-vs-background difference; \(\Dclean\): cross-teacher dispersion; \(H_w\): teacher-credibility entropy. The \(p\)-values are from permutation tests.
\end{flushleft}
\end{table}

\subsection{Spatial Robustness under Off-Center Anchors}
\label{sec:dual-offcenter}

Beyond standard center anchors, we evaluate whether the student remains
stable when the target box is deliberately shifted away from the image
center. Table~\ref{tab:dual_offcenter} reports the dual-domain
off-center anchor perturbation results.

\begin{table}[htbp]
\centering
\small
\caption{Dual-domain off-center anchor perturbation results (\(\Npatch \le 7\)).}
\label{tab:dual_offcenter}
\begin{tabular}{llccccc}
\toprule
Dataset & Level & Offset & \(\Npatch\) & PLCC & \(\Delta\)PLCC & \(r\) \\
\midrule
CSIQ & Center & 0.010 & 50 & \textbf{0.9009} & --- & --- \\
     & Mild & 0.055 & 7 & 0.8815 & $-0.0194$ & 0.9846 \\
     & Moderate & 0.096 & 7 & 0.8838 & $-0.0171$ & 0.9847 \\
     & Severe & 0.115 & 5 & 0.8846 & $-0.0163$ & 0.9851 \\
\midrule
LIVEC & Center & 0.010 & 50 & \textbf{0.9133} & --- & --- \\
      & Mild & 0.052 & 6 & 0.8777 & $-0.0356$ & 0.9535 \\
      & Moderate & 0.080 & 6 & 0.8862 & $-0.0271$ & 0.9525 \\
      & Severe & 0.107 & 6 & 0.8837 & $-0.0296$ & 0.9524 \\
\bottomrule
\end{tabular}

\vspace{2mm}
\begin{flushleft}
\footnotesize \textit{Note:} \(r\) denotes the per-image Pearson correlation between the predicted and ground-truth scores. \(\Delta\)PLCC is the drop relative to the center anchor.
\end{flushleft}
\end{table}

\noindent\textbf{Spatial robustness.} Despite the disjoint distortion
manifolds, the two datasets yield consistent conclusions. Under extreme
offsets (0.115 on CSIQ, 0.107 on LIVEC), PLCC degradation stays strictly
within 3.6\% (CSIQ: $\Delta$PLCC $\le -0.019$; LIVEC: $\Delta$PLCC
$\le -0.036$), and per-image correlation remains exceptionally high
($r \ge 0.952$). The non-monotonic recovery at the severe level is
consistent with the information-coverage effect analyzed in
Section~\ref{sec:ana-delta}, not with a simple monotonic degradation
with offset.

\noindent\textbf{Length-matched control.} Truncating the LIVEC center
sequence to $\Npatch=6$ yields a PLCC of 0.8603; the same-length
off-center sequence (Severe: 0.8837) slightly outperforms it, as larger
spatial coverage provides richer background gradients. Subset length is
therefore ruled out as the source of the observed degradation.

\subsection{Cross-Dataset Transfer}
\label{sec:res-cross}

The cross-dataset setting (L2) measures the zero-shot transferability
of the learned representation. Table~\ref{tab:res-cross} reports the
cross-dataset transfer results. When trained on CSIQ and tested on
LIVEC, the student achieves a PLCC of 0.8416, which exceeds the
consensus (0.6469) by $+0.1947$. When trained on LIVEC and tested on
CSIQ, the student achieves a PLCC of 0.7422, which is essentially equal
to the consensus (0.7427), with $\delta_{\text{PLCC}} = -0.0005$.

\begin{table}[htbp]
\centering
\small
\caption{Cross-dataset transfer results.}
\label{tab:res-cross}
\begin{tabular}{lcccc}
\toprule
Train \(\to\) Test & PLCC & SRCC & KRCC & MAE \\
\midrule
CSIQ \(\to\) CSIQ & 0.9009 & 0.8664 & 0.6825 & 0.0891 \\
\textbf{CSIQ \(\to\) LIVEC} & \textbf{0.8416} & 0.8204 & 0.6283 & 0.0938 \\
LIVEC \(\to\) LIVEC & 0.9133 & 0.8865 & 0.7101 & 0.0719 \\
\textbf{LIVEC \(\to\) CSIQ} & \textbf{0.7422} & 0.7438 & 0.5355 & 0.1386 \\
\bottomrule
\end{tabular}

\vspace{2mm}
\begin{flushleft}
\footnotesize \textit{Note:} All student models are frozen at their source-domain weights (zero-shot) during the target-domain evaluation.
\end{flushleft}
\end{table}

\noindent\textbf{Asymmetry and the zero-shot advantage.} The transfer
results are asymmetric. The CSIQ $\to$ LIVEC transfer succeeds, with
the student substantially exceeding the consensus. Conversely, the
LIVEC $\to$ CSIQ transfer exhibits a relative degradation, with the
student essentially matching the consensus.

A fundamental asymmetry in the evaluation protocol must be noted: the
teacher models (e.g., MUSIQ) are evaluated with their official weights,
which likely overlap with the target datasets, giving them an inherent
domain-adapted advantage. Under the CSIQ $\to$ LIVEC transfer, our
strictly zero-shot student (PLCC 0.8416) closely matches the best
single teacher (MANIQA, PLCC 0.8443) and completely surpasses the
consensus (0.6469). Under the reverse transfer (LIVEC $\to$ CSIQ), the
student achieves a PLCC of 0.7422. Although this is slightly below the
best teacher (MUSIQ, 0.7715), it still matches the consensus baseline
(0.7427) and its SRCC (0.7438) is substantially higher than that of
the consensus under the same transfer (0.5355).

We \emph{hypothesize} that the asymmetry reflects a difference in
source-manifold coverage rather than in dataset size. CSIQ covers a
narrow, controlled set of six synthetic distortion types, whereas LIVEC
covers a broader, uncontrolled set of authentic distortions.
Transferring from CSIQ to LIVEC is a move from a narrower to a broader
distortion manifold, in which the representation learned on the source
still carries transferable structure. The reverse direction is a move
from a broader to a narrower manifold: the source does not expose the
learner to the specific controlled distortions of CSIQ, so the
representation has little additional distortion-type coverage to
contribute. This hypothesis is consistent with the observed pattern but
is not isolated from other factors (dataset size, MOS scale) in the
present experiments; a controlled study that fixes size while varying
coverage remains future work (Section~\ref{sec:concl-limitations}).

Under CSIQ-to-LIVEC, the student both gains globally ($+0.1947$) and
differs from consensus along physically interpretable axes. Under
LIVEC-to-CSIQ, the global PLCC is flat, yet the subgroup analysis in
Appendix~\ref{app:grouped} shows that on the low-$\Dclean$ subgroup---where consensus
collapses to $0.1325$---the student still achieves $0.7896$. This is
consistent with the information-coverage hypothesis: the source
manifold does not cover the target's distortion types, so the student
cannot rely on a transferred prior and must instead exploit
within-image structure that remains available in the target domain. A
flat aggregate therefore reflects information complementarity under a
strong domain-adapted baseline, not a failure of the differencing
mechanism.

\subsection{Ablation Studies}
\label{sec:res-ablation}

The ablation configurations (L3) each disable one specific module of
the student network. Table~\ref{tab:ablation-config} lists the
configurations and the operation used to disable each module. The
ablation is performed by freezing the trained model and modifying the
forward pass at inference time; no retraining is involved. This
protocol isolates the direct contribution of each module under
identical initial conditions.

\begin{table}[htbp]
\centering
\small
\caption{Ablation configurations.}
\label{tab:ablation-config}
\begin{tabular}{cll}
\toprule
ID & Configuration & Disabling operation \\
\midrule
A0 & Full model & --- \\
A1 & No sequence encoder & \(\zseq = 0\) \\
A2 & No geometric gate & \(g_k^{\text{geo}} = 1/\Npatch\) \\
A3 & No dispersion gate & \(\tilde g_k^{\text{dis}} = 1\) \\
A4 & No both gates & \(\gagg_k = 1/\Npatch\) \\
A5 & No per-teacher scale & \(\seff_i = 1\) \\
A6 & No scale correction & \(\eta_i = 0\) \\
A7 & No background correction & \(\hat\beta_i = 0\) \\
\bottomrule
\end{tabular}

\vspace{2mm}
\begin{flushleft}
\footnotesize \textit{Note:} A0 is the full model; A1--A7 each disable one module at inference time; no retraining is involved.
\end{flushleft}
\end{table}

Table~\ref{tab:res-ablation-all} reports the PLCC and MAE for all
configurations across the three datasets.

\begin{table}[htbp]
\centering
\small
\caption{Ablation results on the three datasets.}
\label{tab:res-ablation-all}
\setlength{\tabcolsep}{3pt} 
\begin{tabular}{lcccccc}
\toprule
 & \multicolumn{2}{c}{Self-built} & \multicolumn{2}{c}{CSIQ} & \multicolumn{2}{c}{LIVEC} \\
\cmidrule(lr){2-3} \cmidrule(lr){4-5} \cmidrule(lr){6-7}
Config & PLCC & MAE & PLCC & MAE & PLCC & MAE \\
\midrule
A0 (full) & \textbf{0.9855} & \textbf{0.0198} & \textbf{0.9009} & \textbf{0.0891} & \textbf{0.9133} & \textbf{0.0719} \\
A1 (no seq.) & 0.8421 & 0.4437 & 0.7415 & 0.1429 & 0.8145 & 0.1023 \\
A2 (no geo.) & 0.9331 & 0.0431 & 0.8792 & 0.0968 & 0.9091 & 0.0733 \\
A3 (no disp.) & 0.9855 & 0.0198 & 0.9009 & 0.0891 & 0.9133 & 0.0719 \\
A4 (no both) & 0.9332 & 0.0431 & 0.8792 & 0.0968 & 0.9090 & 0.0733 \\
A5 (no scale) & 0.2475 & 39.2301 & 0.7612 & 0.1387 & 0.7557 & 0.1168 \\
A6 (no corr.) & 0.9855 & 0.0198 & 0.6358 & 0.1687 & 0.7362 & 0.1123 \\
A7 (no bg.) & 0.8604 & 0.3140 & 0.8615 & 0.1056 & 0.8640 & 0.0889 \\
\bottomrule
\end{tabular}

\vspace{1mm}
\begin{flushleft}
\footnotesize \textit{Note:} A1: no sequence encoder; A2: no geometric gate; A3: no dispersion gate; A4: no both gates; A5: no per-teacher scale; A6: no scale correction; A7: no background correction.
\end{flushleft}
\end{table}

\noindent\textbf{Module importance.} Table~\ref{tab:res-module-rank}
summarizes the module ranking by mean $\delta_{\text{PLCC}}$ over the
three datasets. Per-teacher scale (A5), scale correction (A6), and
sequence encoder (A1) are the most important modules; background
correction (A7) has moderate importance; geometric gate (A2/A4) has
small importance; dispersion gate (A3) is accuracy-redundant ($\Delta
= 0.0000$ on all three datasets). The A5--A6 ranking is
dataset-dependent: A5 dominates on the self-built dataset (large scale
mismatch), while A6 dominates on CSIQ and LIVEC (base scales already
near optimal). The mean over the three datasets places A5 first,
consistent with the large scale mismatch of the 11-teacher ensemble.

\begin{table}[htbp]
\centering
\small
\caption{Module importance ranking across the three datasets.}
\label{tab:res-module-rank}
\begin{tabular}{clcccc}
\toprule
Rank & Module & Self-built & CSIQ & LIVEC & Mean \\
\midrule
1 & Per-teacher scale (A5) & \(-0.7380\) & \(-0.1397\) & \(-0.1576\) & \(-0.3451\) \\
2 & Scale correction (A6) & \(0.0000\) & \(-0.2649\) & \(-0.1771\) & \(-0.1473\) \\
3 & Sequence encoder (A1) & \(-0.1434\) & \(-0.1594\) & \(-0.0988\) & \(-0.1339\) \\
4 & Background correction (A7) & \(-0.1251\) & \(-0.0394\) & \(-0.0493\) & \(-0.0713\) \\
5 & Geometric gate (A2/A4) & \(-0.0524\) & \(-0.0217\) & \(-0.0042\) & \(-0.0261\) \\
6 & Dispersion gate (A3) & \(0.0000\) & \(0.0000\) & \(0.0000\) & \(0.0000\) \\
\bottomrule
\end{tabular}

\vspace{1mm}
\begin{flushleft}
\footnotesize \textit{Note:} Sorted by mean \(\delta_{\text{PLCC}}\). A6 has no effect on the self-built dataset because the learned log-scale correction is already near zero. A5 has the largest drop because the self-built teachers are on very different numerical scales.
\end{flushleft}
\end{table}

\noindent\textbf{The diagnostic role of A3.} A3 (the dispersion gate)
leaves PLCC unchanged under center anchors ($\Delta=0.0000$), which
might suggest it could be removed. However, A3 serves a dual role. In
terms of accuracy, it is a passive reliability filter that does not
affect the final prediction under well-centered anchors. In terms of
diagnostics, the statistic $d_k^{\text{dis}}$ that feeds A3 is
precisely the quantity used to compute \(\Dclean\) in the $\Delta$
analysis (Section~\ref{sec:ana-delta}). Removing A3 from the
architecture would sever this diagnostic pathway and reduce the
residual analysis to a purely post-hoc statistical exercise. We
therefore retain A3 as an architectural interface for interpretability,
while acknowledging that its accuracy contribution is nominal. Designing
an active dispersion gate that genuinely improves accuracy under
extreme shifts remains an open problem.

\noindent\textbf{Note on A5.} Removing the per-teacher scale (A5) causes
a catastrophic failure on the self-built dataset
(MAE \(= 39.2301\)). This reflects a numerical-scale mismatch rather
than a structural failure. The raw teacher readings span a wide
numerical range (e.g., PaQ-2-PiQ is on the order of 70, whereas
CLIP-IQA+ is below 1), while the weak label generated by the consensus
is normalized to a small range. Without \(s_i^{\text{eff}}\), the
unnormalized input pushes the predicted \(\hat Q\) to a scale of tens,
and the resulting MSE gradient diverges during training. On CSIQ and
LIVEC, where the ground-truth MOS is already on a larger scale, A5
degrades gracefully (MAE \(\approx 0.13\)), confirming that the
catastrophic failure is specific to the target scale of the self-built
dataset. A monotone rescaling of each teacher would not change PLCC,
but the relative weighting across teachers is what \(s_i^{\text{eff}}\)
controls; removing it therefore destroys the numerical conditioning of
the network input. We intentionally did not clamp the output because
the failure itself demonstrates the necessity of \(s_i^{\text{eff}}\)
for numerical stability.

\subsection{\texorpdfstring{$\Delta$}{Delta} Structure Analysis}
\label{sec:ana-delta}

The ablation established which modules are necessary; the $\Delta$
analysis addresses a stronger question: whether the student learns
information that the consensus does not contain. This is the central
test of the learning claim, because $\Delta = \hat{Q} -
\mathrm{consensus}$ is computed against human MOS on both benchmarks,
and a student that merely reweights the teachers would produce a
$\Delta$ uncorrelated with any physical quantity. The formal definition
of $\Delta$ and the candidate quantities are given in
Section~\ref{sec:delta}. Here, we report the empirical results by
analyzing the residual $\Delta = \hat{Q} - \mathrm{consensus}$ and its
correlation with five physically interpretable quantities:
$\Dtgtbg$, $\Dclean$, $\Cintra$, $\Ewl$, and $\Sarea$. The $\Delta$
analysis is reported on CSIQ and LIVEC, the two benchmarks with human
MOS; on the self-built dataset no target-level MOS exists, so $\Delta$
is interpreted against geometric and dispersion quantities in
Section~\ref{sec:res-within}.

Tables~\ref{tab:delta-csiq} and~\ref{tab:delta-livec} report the
results on CSIQ and LIVEC after training. On CSIQ, the differences on
$\Dclean$, $\Cintra$, and $\Ewl$ are significant; on LIVEC, all five
quantities are significantly different from zero. The $\Dclean$ row is
the most informative: consensus error is $-0.8834$ against MOS, while
the student error is $-0.0770$, i.e., consensus is almost
anti-correlated with MOS on high-dispersion samples, whereas the
student is not. This is the signature of learning beyond the ensemble,
not of reweighting it.

\begin{table}[htbp]
\centering
\small
\caption{\(\Delta\) structure analysis on CSIQ after training.}
\label{tab:delta-csiq}
\begin{tabular}{lcccc}
\toprule
Quantity & BiCG-Net error & Consensus error & Fisher \(z\) & \(p\) \\
\midrule
\(\Dtgtbg\) & \(+0.0660\) & \(+0.0567\) & \(0.193\) & \(0.847\) \\
\(\Dclean\) & \(-0.0770\) & \(-0.8834\) & \(27.296\) & \(0.00\) \\
\(\Cintra\) & \(+0.1672\) & \(+0.3064\) & \(-3.071\) & \(2.1\times10^{-3}\) \\
\(\Ewl\) & \(-0.0508\) & \(-0.3347\) & \(6.174\) & \(6.7\times10^{-10}\) \\
\(\Sarea\) & \(-0.1627\) & \(-0.2469\) & \(1.828\) & \(0.068\) \\
\bottomrule
\end{tabular}

\vspace{2mm}
\begin{flushleft}
\footnotesize \textit{Note:} \emph{Error} = Pearson correlation between the estimator's deviation from MOS (\(\text{estimator} - \text{MOS}\)) and the corresponding physical quantity; a negative value means the deviation is anti-correlated with that quantity.
\end{flushleft}
\end{table}

\begin{table}[htbp]
\centering
\small
\caption{\(\Delta\) structure analysis on LIVEC after training.}
\label{tab:delta-livec}
\begin{tabular}{lcccc}
\toprule
Quantity & BiCG-Net error & Consensus error & Fisher \(z\) & \(p\) \\
\midrule
\(\Dtgtbg\) & \(+0.0705\) & \(+0.1863\) & \(-2.837\) & \(4.6\times10^{-3}\) \\
\(\Dclean\) & \(-0.1946\) & \(-0.6938\) & \(15.843\) & \(0.00\) \\
\(\Cintra\) & \(+0.0179\) & \(+0.1869\) & \(-4.122\) & \(3.8\times10^{-5}\) \\
\(\Ewl\) & \(-0.0628\) & \(-0.1976\) & \(3.308\) & \(9.4\times10^{-4}\) \\
\(\Sarea\) & \(-0.1117\) & \(-0.3380\) & \(5.769\) & \(8.0\times10^{-9}\) \\
\bottomrule
\end{tabular}

\vspace{2mm}
\begin{flushleft}
\footnotesize \textit{Note:} \emph{Error} = Pearson correlation between the estimator's deviation from MOS (\(\text{estimator} - \text{MOS}\)) and the corresponding physical quantity.
\end{flushleft}
\end{table}

\noindent\textbf{Summary of $\Delta$ analysis.} This is why the
student exceeds the teachers: the residual $\Delta$ is significantly
correlated with several physically interpretable quantities, showing
that the student is not a re-weighting of teacher outputs. The pattern
differs: on CSIQ, the student differs from consensus mainly on
dispersion, consistency, and extremity; on LIVEC, the student differs
on all five quantities. The significant correlations indicate that the
residual carries information beyond consensus, as predicted by
Proposition~\ref{prop:delta}. This confirms that BiCG-Net captures
background-aware cues and teacher dispersion patterns that a simple
weighted average omits.
\section{Conclusion}
\label{sec:conclusion}

\subsection{Summary of Findings}
\label{sec:concl-summary}

This work addressed target-level image quality assessment without
target-level human MOS. We formalized the teacher reading as an
underdetermined observation model, proved that in-image differencing
exactly cancels the target term and the teacher-specific bias, and
designed BiCG-Net as a compact student network that aggregates the
resulting intra-image statistics. On the two MOS benchmarks, BiCG-Net
achieves PLCCs of 0.9009 (CSIQ) and 0.9133 (LIVEC), both outperforming
the best single teacher and the teacher consensus against human MOS. On
a self-built MOS-free railway dataset, it achieves a PLCC of 0.9855
against the weak label. Under strictly zero-shot CSIQ-to-LIVEC
transfer, the frozen student reaches a PLCC of 0.8416, exceeding the
consensus by $+0.1947$.

The key contributions are threefold: a nested-sampling formulation with provable bias elimination; an interpretable four-module architecture with a bounded objective; and a dual-track validation with a residual diagnostic. Two findings deserve emphasis. First, a distilled student
can exceed its teachers because it is not trained to mimic any
individual teacher, but to predict the ground-truth label from the
\emph{relative structure} of the teacher readings; a residual analysis
confirms that this departure from consensus is significantly correlated
with physically interpretable quantities. Second, the dispersion gate
(A3) does not contribute to accuracy ($\Delta$PLCC $= 0.0000$ across
all three datasets); it is a diagnostic module whose statistic
$d_k^{\text{dis}}$ feeds the residual analysis. We therefore position
BiCG-Net as an offline target-level quality auditing and
label-generation engine: its 39,482-parameter student is compact by
design, and the pipeline bottleneck resides in the teacher ensemble.

\subsection{Limitations and Future Work}
\label{sec:concl-limitations}

Several limitations suggest directions for future work.

\noindent\textbf{The diagnostic role of A3.} The dispersion gate
(A3) leaves PLCC unchanged under center anchors, and the off-center
analysis showed that teacher disagreement is not monotonically driven
by off-centering. A3 is therefore a diagnostic module rather than an
accuracy-oriented one. Designing a gating mechanism that improves
accuracy under extreme spatial shifts remains an open question.

\noindent\textbf{Alignment with human perception on the self-built dataset.}
While the $\Delta$ analysis (Table~\ref{tab:delta-self}) confirms that
the student captures structural information beyond a simple consensus
average, the 0.9855 PLCC is measured against the weak label itself. It
therefore reflects the model's fidelity to the weak supervision, not
direct agreement with human perception. Future work will introduce a
subset of human MOS annotations for the self-built dataset to measure
the correlation between the weak label and human judgment, thereby
establishing the absolute perceptual ceiling of this track.

\noindent\textbf{Asymmetric transfer.} Cross-dataset transfer is
asymmetric: CSIQ~$\to$~LIVEC succeeds, but LIVEC~$\to$~CSIQ is
essentially flat. We hypothesize an information-coverage asymmetry, but
this is not isolated from other factors (dataset size, MOS scale) in
the present experiments. A controlled study that fixes dataset size
while varying distortion coverage remains future work.

\noindent\textbf{Sparse off-center anchors.} Due to randomized jitter
in the fixed sampling template, legally nested anchors beyond offset
0.115 are extremely sparse. Future work will regenerate strictly nested
patches from original images to probe larger offsets without
sacrificing subset density.

\noindent\textbf{Extensions.} The framework can be extended to
multi-object scenarios within the same frame, and to other industrial
pipelines (e.g., power-line diagnostics and wind-turbine blade
monitoring) with small target coverage and no target-level annotation.
Empirical validation in these settings remains future work.

\section*{Acknowledgments}
This work was supported by CRRC Chongqing Co., Ltd., under the project
titled ``Development of an Artificial Intelligence Large Model for
Urban Transportation Vehicle Exterior Styling and Industrial Design''
(Project No.~2024CCA572). The authors gratefully acknowledge the
comprehensive financial and technical support provided for this
research.

\section*{Data and Code Availability}
The source code and dataset generated and analyzed during the current
study are publicly available in the GitHub repository
\url{https://github.com/cqugege/BiCG-Net}.

\section*{Conflict of Interest}
The authors declare that they have no known competing financial
interests or personal relationships that could have appeared to
influence the work reported in this paper.

\appendix

\section{Teacher Correlation Matrix}
\label{app:teacher-correlation}

\begin{figure}[htbp]
  \centering
  \includegraphics[width=0.7\linewidth]{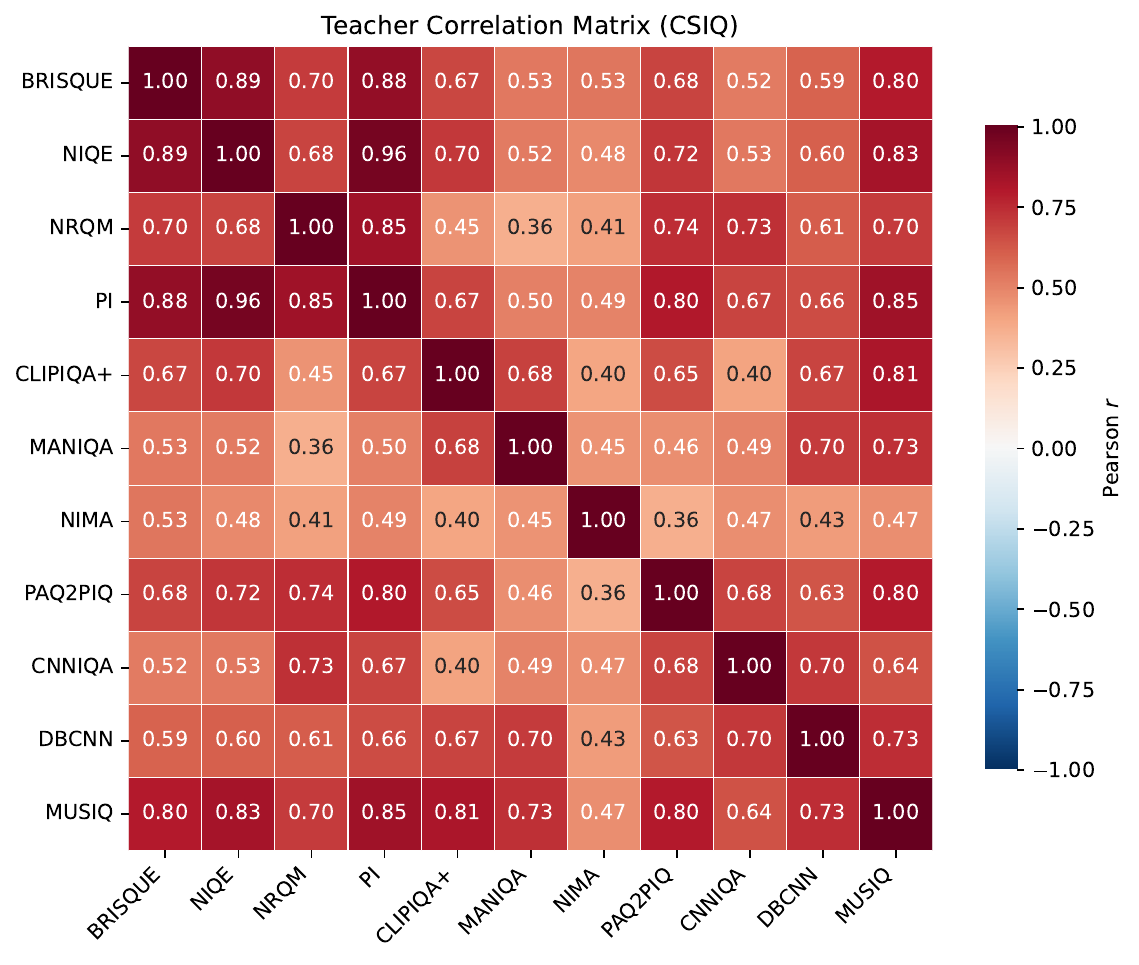}
  \caption{Pairwise Pearson correlation among the 11 teachers on CSIQ ($n = 866$).}
  \label{fig:teacher-corr-csiq}
\end{figure}

\begin{figure}[htbp]
  \centering
  \includegraphics[width=0.7\linewidth]{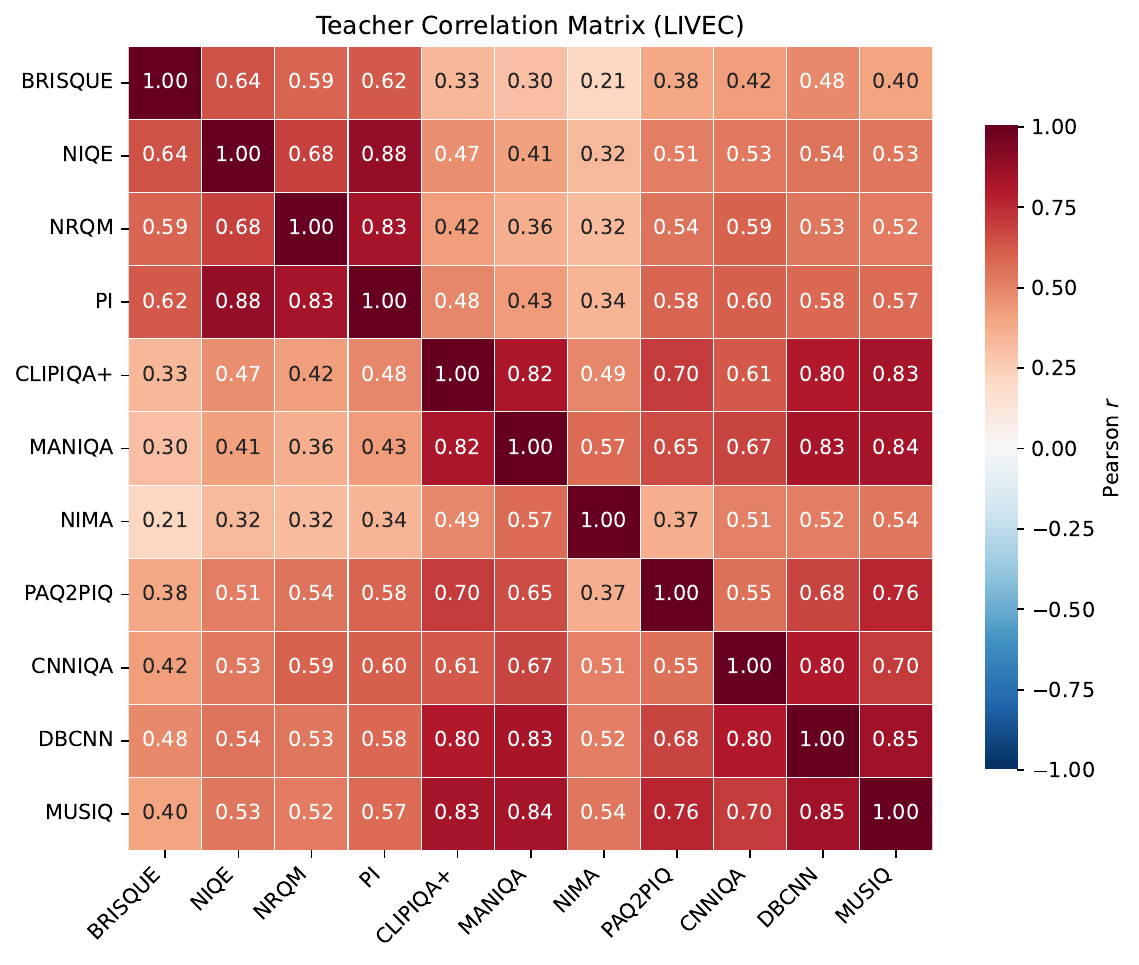}
  \caption{Pairwise Pearson correlation among the 11 teachers on LIVEC ($n = 1,162$).}
  \label{fig:teacher-corr-livec}
\end{figure}

Figure~\ref{fig:teacher-corr-csiq} and Figure~\ref{fig:teacher-corr-livec}
report the pairwise Pearson correlation among the 11 teachers on CSIQ
and LIVEC, respectively. On CSIQ ($n = 866$), the correlations range
from $r = 0.363$ to $r = 0.955$ (median $r = 0.667$); on LIVEC
($n = 1,162$), they range from $r = 0.215$ to $r = 0.881$ (median
$r = 0.544$). In both datasets, no pair reaches $r = 1$, indicating
that the 11 teachers are not redundant and that the ensemble covers
complementary aspects of image quality. The NIQE--PI pair is the most
redundant across both datasets, whereas the deep-learning teachers
(e.g., NIMA, MANIQA) are weakly correlated with the traditional ones on
LIVEC ($r$ as low as $0.215$). This observation justifies our choice of
combining both families in the ensemble.

\section{Grouped Analysis}
\label{app:grouped}

Table~\ref{tab:grouped-all} reports the PLCC of BiCG-Net and the
consensus on subgroups defined by the tertiles of $\Dtgtbg$ and
$\Dclean$, as defined in Section~\ref{sec:delta}. The student network
remains stable across all subgroups, whereas the consensus collapses on
the low-$\Dclean$ subgroup of CSIQ (PLCC $= 0.1325$), where BiCG-Net
still achieves $0.7896$. This highlights the robustness of our method
in cases where teacher disagreements are minimal.

\begin{table}[htbp]
\centering
\small
\caption{Grouped PLCC on CSIQ and LIVEC (after training).}
\label{tab:grouped-all}
\begin{tabular}{llcccc}
\toprule
 & & \multicolumn{2}{c}{CSIQ} & \multicolumn{2}{c}{LIVEC} \\
\cmidrule(lr){3-4} \cmidrule(lr){5-6}
Grouping & Subgroup & BiCG-Net & Consensus & BiCG-Net & Consensus \\
\midrule
\multirow{3}{*}{\(\Dtgtbg\)}
  & Low  & 0.8986 & 0.7530 & 0.9032 & 0.4940 \\
  & Mid  & 0.8696 & 0.7247 & 0.8876 & 0.6565 \\
  & High & 0.9102 & 0.7460 & 0.9367 & 0.7041 \\
\midrule
\multirow{3}{*}{\(\Dclean\)}
  & Low  & 0.7896 & 0.1325 & 0.8855 & 0.5984 \\
  & Mid  & 0.7606 & 0.4055 & 0.9133 & 0.7552 \\
  & High & 0.9105 & 0.8043 & 0.9353 & 0.6895 \\
\bottomrule
\end{tabular}

\vspace{2mm}
\begin{flushleft}
\footnotesize \textit{Note:} Subgroups are defined by the tertiles of \(\Dtgtbg\) and \(\Dclean\) (see Section~\ref{sec:delta}).
\end{flushleft}
\end{table}

\section{Computational Cost Breakdown}
\label{app:cost}

Table~\ref{tab:cost-breakdown} reports the per-stage latency
breakdown of the full pipeline. The 11-teacher IQA scoring stage
dominates the runtime, accounting for 99.96\% of the total
pipeline time, whereas the proposed student network adds only
$0.0013$\,s ($0.0001\%$ of the total).

\begin{table}[htbp]
\centering
\small
\caption{Per-stage latency breakdown. The 11-teacher scoring dominates
the pipeline. The 294 images are processed on 2 GPUs; timings are
per-image wall-clock.}
\label{tab:cost-breakdown}
\begin{tabular}{lccc}
\toprule
Stage & Mean (s) & Std (s) & Share (\%) \\
\midrule
Target Detection (YOLO)        & 0.04095 & 0.01571 & 0.00325 \\
Fine Segmentation (SAM)        & 0.44472 & 0.06726 & 0.03533 \\
Topological Analysis           & 0.01734 & 0.02841 & 0.00138 \\
Patch Sampling                 & 0.000338 & 0.0000937 & 0.0000269 \\
Baseline 11-IQA Scoring        & 1258.248 & 1101.410 & 99.9599 \\
BiCG-Net Inference             & 0.001335 & 0.001112 & 0.000106 \\
\midrule
Total Pipeline                 & 1258.753 & 1101.442 & 100.0 \\
\bottomrule
\end{tabular}
\end{table}

Table~\ref{tab:teacher-cost} further decomposes the 11-teacher cost.
NRQM and PI alone account for 83.1\% of the total teacher time.

\begin{table}[htbp]
\centering
\small
\caption{Per-teacher inference cost, aggregated over the 294 evaluation samples.}
\label{tab:teacher-cost}
\begin{tabular}{lcccc}
\toprule
Teacher & Mean (s) & Std (s) & Total over 294 samples (s) & Share (\%) \\
\midrule
BRISQUE   & 5.545   & 0.869   & 1630.30   & 0.441 \\
CLIP-IQA+ & 7.636   & 6.267   & 2244.87   & 0.607 \\
MANIQA    & 130.362 & 13.751  & 38326.29  & 10.361 \\
NIQE      & 5.995   & 1.121   & 1762.46   & 0.476 \\
NIMA      & 8.374   & 0.924   & 2461.86   & 0.666 \\
PaQ-2-PiQ & 1.816   & 1.002   & 533.88    & 0.144 \\
CNNIQA    & 38.064  & 62.737  & 11190.72  & 3.025 \\
DBCNN     & 6.193   & 9.218   & 1820.69   & 0.492 \\
MUSIQ     & 8.604   & 19.099  & 2529.53   & 0.684 \\
NRQM      & 521.227 & 496.012 & 153240.69 & 41.425 \\
PI        & 524.435 & 503.494 & 154183.76 & 41.680 \\
\bottomrule
\end{tabular}
\end{table}

\clearpage
\bibliographystyle{elsarticle-num}
\bibliography{refs}

\end{document}